\documentclass{article}

 \usepackage[preprint]{neurips_2026}

\usepackage[utf8]{inputenc} 
\usepackage[T1]{fontenc}    
\usepackage{hyperref}       
\hypersetup{hidelinks}
\usepackage{url}            
\usepackage{booktabs}       
\usepackage{amsfonts}       
\usepackage{amsmath}        
\usepackage{nicefrac}       
\usepackage{microtype}      
\usepackage[table]{xcolor}   
\usepackage{graphicx}
\usepackage{enumitem}       
\usepackage{amssymb}  
\usepackage{multirow}
\usepackage{subcaption}       
\usepackage{wrapfig}          
\usepackage{dblfloatfix}      
\usepackage{placeins}         
\usepackage{xspace}
\usepackage{bm}

\definecolor{oursrow}{RGB}{218,232,252}   
\definecolor{grouphdr}{RGB}{234,234,234}  
\definecolor{bestcell}{RGB}{255,243,205}  
\definecolor{matchhdr}{RGB}{226,239,218}  
\definecolor{gaincell}{RGB}{226,239,218}  
\definecolor{losscell}{RGB}{244,204,204}  

\title{LR-V2X: Loss-Resilient Collaborative Perception under Low-Bandwidth Communication}
\setcitestyle{numbers,square,comma}
\newcommand{\methodname}{LR-V2X\xspace}
\makeatletter
\renewcommand{\@noticestring}{Accepted at NeurIPS 2026.}
\makeatother

\author{%
  {\bfseries Kang Yang$^1$ \quad Tianci Bu$^2$ \quad Peng Wang$^1$} \\
  {\bfseries Deying Li$^1$ \quad Yongcai Wang$^{1,3}$\thanks{Corresponding author.}} \\[0.5em]
  {\small $^1$Renmin University of China} \\
  {\small $^2$The Hong Kong University of Science and Technology (HKUST)} \\
  {\small $^3$Hebei Key Laboratory of Real-virtual Integrated} \\
  {\small Autonomous Systems (RIAS)} \\
  {\small \texttt{y1127238112@gmail.com}, \texttt{btc010001@gmail.com}} \\
  {\small \texttt{\{peng.wang,deyingli,ycw\}@ruc.edu.cn}}
}

\begin{document}

\maketitle

\begin{abstract}

Given the inherent unpredictability of packet loss in vehicular wireless communications, V2X collaborative perception can yield practical benefits only if agents can achieve reliable collaboration under lossy and low-bandwidth communication conditions.
Existing dense BEV feature fusion methods depend on redundant BEV feature exchange, which is infeasible in low-bandwidth scenarios, while compact-communication methods aggressively compress messages but can hardly recover the missing feature content after packet loss. 
In this paper, we present \methodname, a loss-resilient, latent-space reconstruction framework that converts corrupted received latents (even under severe 90\% packet loss) into a spatial prior and then reconstructs the missing BEV information from this informative prior and using ego context as condition. Notably, the model can be trained under complete communication conditions and can be directly applied to lossy conditions at test time, eliminating the need for training under numerous lossy conditions.  Experiments on DAIR-V2X and V2XREAL show that \methodname delivers the strongest robustness under severe packet loss and preserves reliable collaboration as communication quality degrades. And it reduces communication overhead by $64\times$ compared to dense BEV feature fusion baselines. Code will be released at \url{https://github.com/sidiangongyuan/LR-V2X}.

\end{abstract}

\begin{figure}[t]
  \centering
  \includegraphics[width=\linewidth]{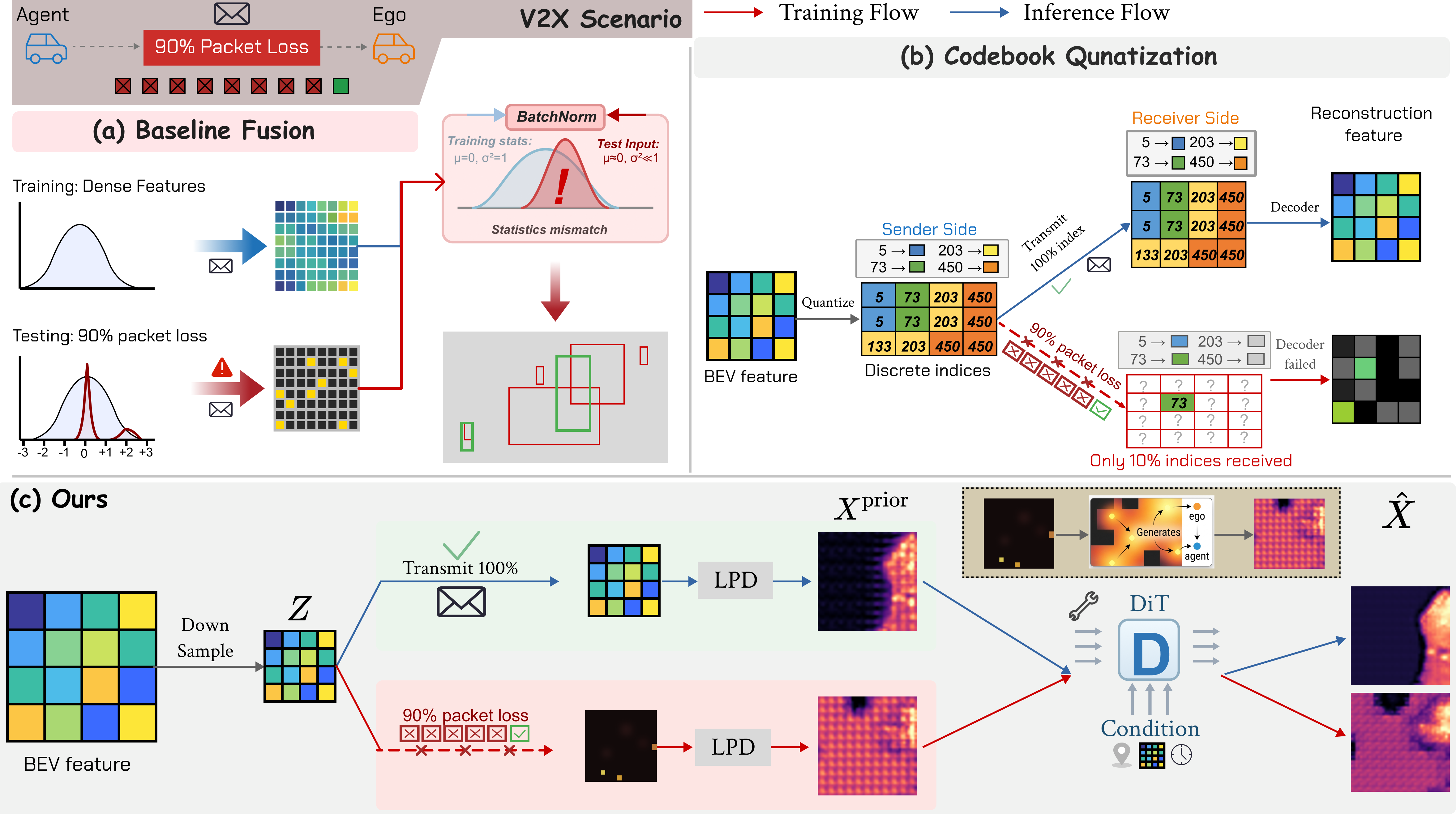}
  \caption{Motivation and overview. (a)~Baseline fusion receives spatially incomplete feature maps when spatial packets are lost. (b)~Codebook-based methods transmit compact codes, but missing indices leave spatial holes that degrade detection. (c)~\methodname reconstructs complete BEV features from corrupted received latents via prior-initialized feature reconstruction, enabling graceful degradation under packet loss.}
  \label{teaser}
\end{figure}

\section{Introduction}
V2X collaborative perception extends the sensing coverage of the ego vehicle by sharing intermediate representations among multiple agents, thereby enhancing driving safety in complex traffic scenarios \cite{OPV2V,dair-v2x,v2v4real,v2xreal}. At present, mainstream intermediate fusion paradigms predominantly transmit BEV feature maps, enabling the ego agent to fuse multi-agent contextual information for accurate 3D object detection \cite{Cooper1,Cooper2,v2vnet,OPV2V,v2x-vit}. However, a critical practical bottleneck lies in maintaining reliable collaborative perception under real-world wireless communication constraints. The \emph{bandwidth limits} force each agent to communicate compact features, while \emph{packet loss} may randomly drop spatial portions of those transmitted features during delivery \cite{collaborative-survey1,collaborative-survey2,collaborative-survey3}. Therefore, under low-bandwidth conditions, the ego agent often receives compact yet incomplete information from collaborative agents. Reliable cooperation thus hinges on effectively recovering meaningful BEV contextual features from such corrupted transmissions. Fig.~\ref{teaser} illustrates this failure scenario and our reconstruction-first paradigm.

Existing methods have only addressed this deployment bottleneck partially. Dense BEV feature fusion baselines \cite{OPV2V,v2x-vit,HEAL,CoBEVT} can preserve accuracy by transmitting highly redundant BEV features, but this robustness is tied to heavy communication overhead. Communication-efficient methods \cite{where2comm,CodeFilling,quantv2x,TransIFF,quest} reduce traffic through region selection, compact codes, or learned compression, yet once packets are lost, they generally lack an explicit mechanism to recover the missing feature content. The central challenge is therefore not only how to communicate less, but how to maintain reliable collaboration if the limited transmitted information is further corrupted during transmission.

\begin{figure}[htbp]
  \centering
  \begin{minipage}[b]{0.32\textwidth}
    \centering
    \includegraphics[width=\linewidth]{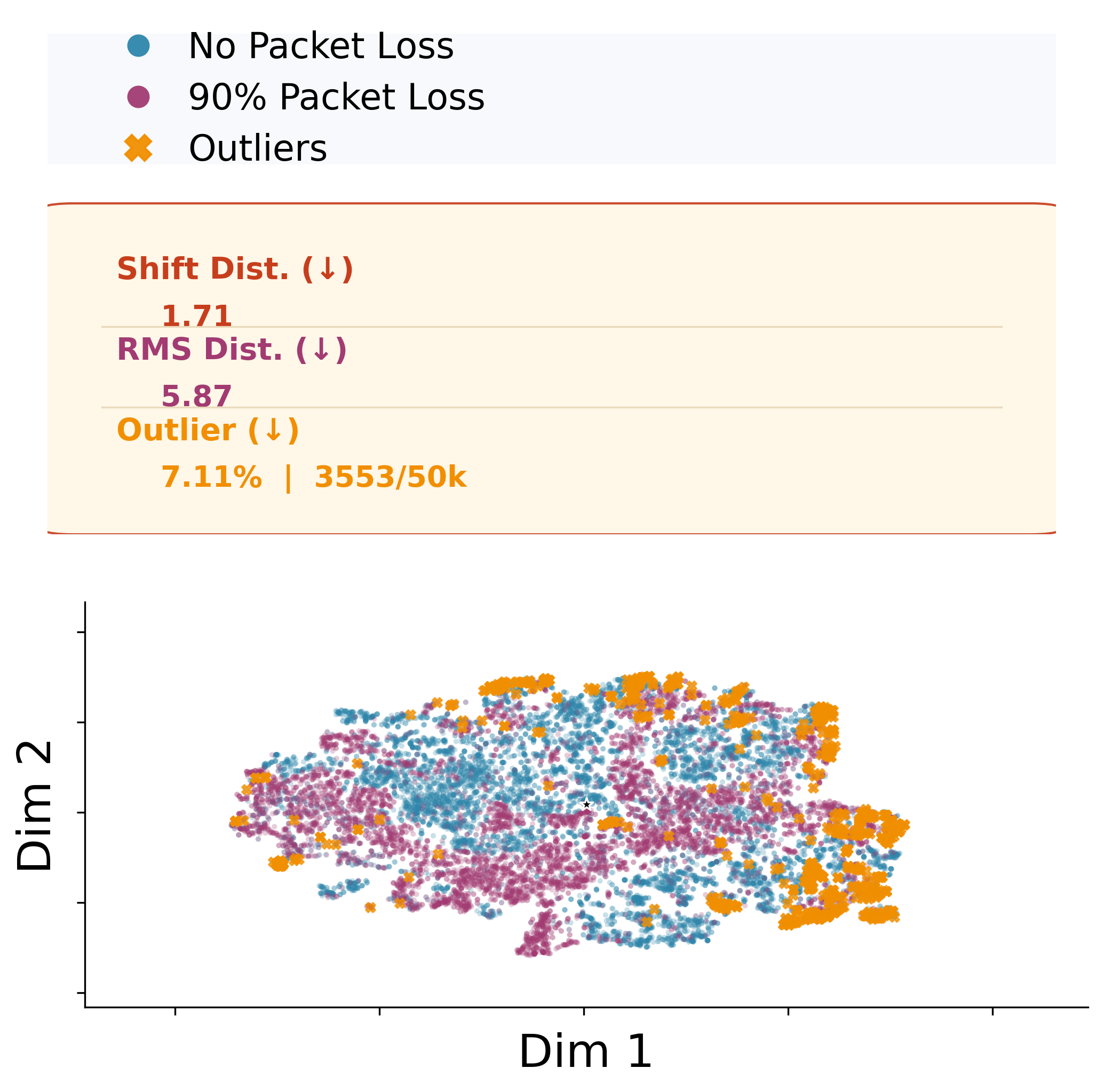}
    \vspace{-0.6em}
    {\scriptsize F-Cooper}
  \end{minipage}
  \hfill
  \begin{minipage}[b]{0.32\textwidth}
    \centering
    \includegraphics[width=\linewidth]{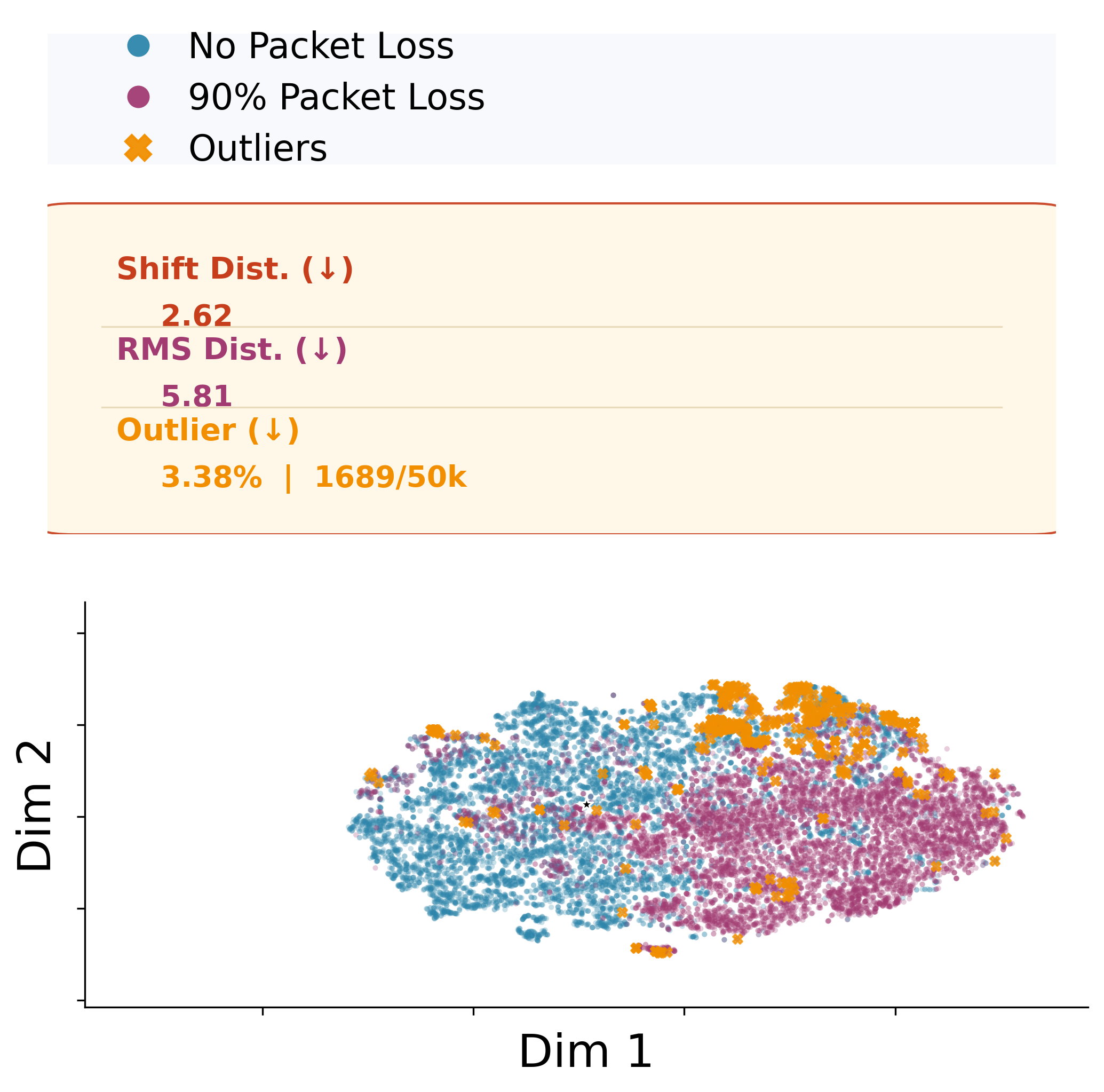}
    \vspace{-0.6em}
    {\scriptsize CodeFilling}
  \end{minipage}
  \hfill
  \begin{minipage}[b]{0.32\textwidth}
    \centering
    \includegraphics[width=\linewidth]{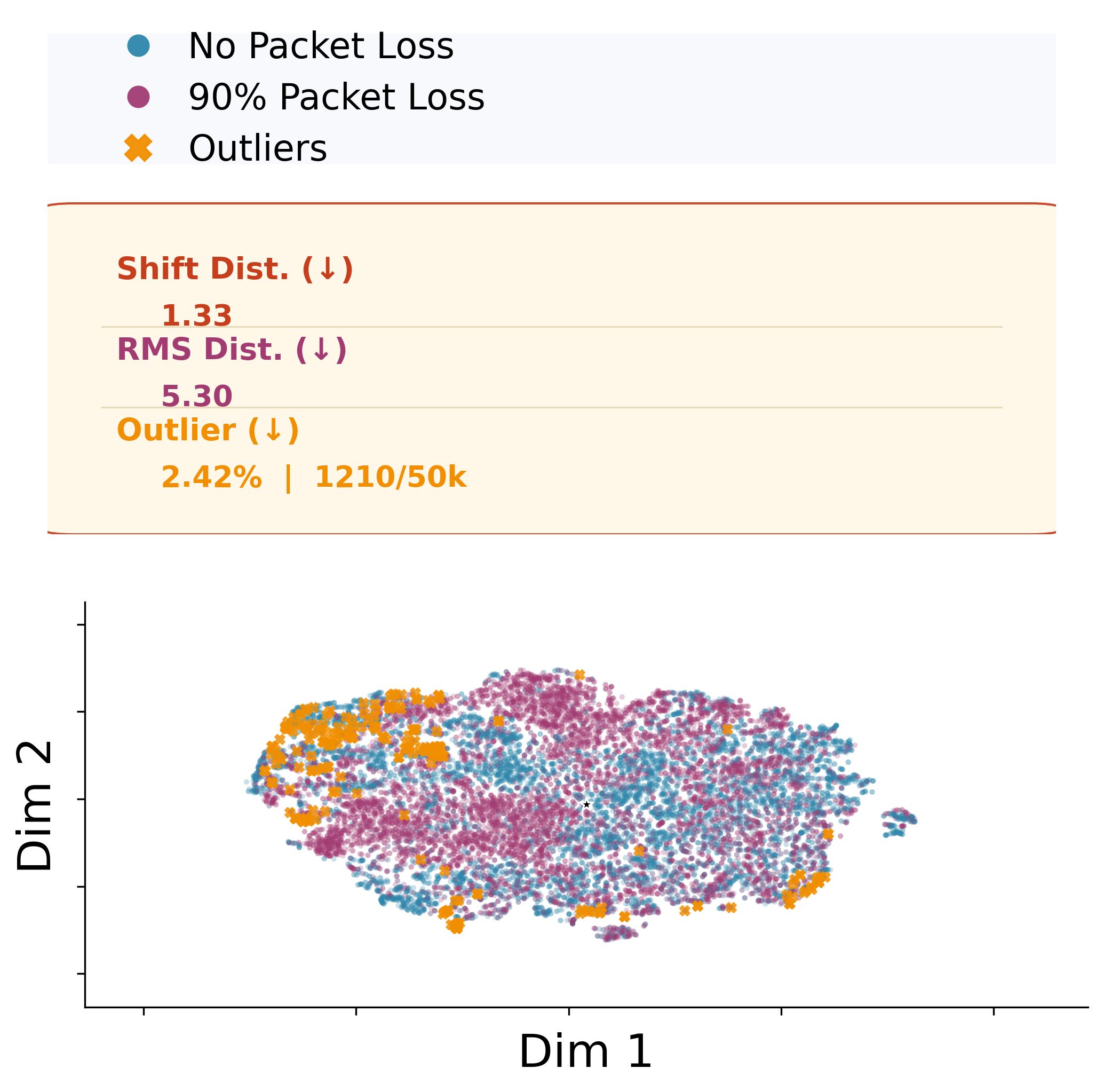}
    \vspace{-0.6em}
    {\scriptsize \methodname}
  \end{minipage}
  \caption{Packet-loss-induced distribution shift in BEV features on DAIR-V2X. Panels show F-Cooper, CodeFilling, and \methodname from left to right. Shift Dist. measures the Mahalanobis distance between packet-loss and reference centroids, RMS Dist. measures the root-mean-square Mahalanobis distance of packet-loss features, and Outlier Rate counts packet-loss features outside the reference high-tail threshold. Lower values indicate a smaller departure from the no-packet-loss reference regime.}
  \label{analysis_of_distribution_shift}
\end{figure}

This problem is difficult because packet loss does not merely remove local evidence. It also induces an input distribution shift absent under ideal communication. When a received feature map becomes spatially sparse, its statistics deviate substantially from those observed during training with complete messages. As shown in Fig.~\ref{analysis_of_distribution_shift}, severe packet loss moves the packet-loss regime away from the no-packet-loss reference regime, so the fusion module must operate on inputs that are systematically mismatched to its training distribution. A dense BEV feature fusion baseline such as F-Cooper can partially buffer this shift through communication redundancy, but in the low-bandwidth regime reliable collaboration requires explicit recovery of missing regions rather than reliance on redundancy alone. This motivates a reconstruction-centric design for reliable low-bandwidth collaboration under packet loss \cite{OOD1,OOD2}.

To reconstruct a neighbor's BEV feature from a received, packet-corrupted compact latent, we propose two key components for loss-resilient V2X: (1) 
inspired by the capability of overlapping transposed convolutions \cite{depth-complete,sparse2dense}, which can propagate evidence from received locations into missing spatial regions, we design 
a \textbf{Latent Prior Decoder (LPD)}. LPD extrapolates features from the successfully received locations into a spatially complete prior BEV map. This prior BEV map guides the following noise-conditioned reconstructor to reconstruct from a reasonable prior, instead of from pure noise.
(2) a \textbf{noise-conditioned DiT \cite{DiT} reconstructor} that reconstructs from the prior BEV map and is guided by ego-side context. The key insight is that the ego-side observation provides supplementary information for reconstructing corrupted latent features received from neighbors, since the ego agent generally shares a large common view with the neighbors. 
Therefore, the noise-conditioned DiT reconstructor refines the prior constructed by LPD, conditioned jointly on the received compact latent and the ego agent's own BEV to fill the missing regions.
Following a diffusion-style training strategy, we perturb the prior with noise at randomly sampled levels, so the denoiser learns to recover from priors of varying quality and generalizes to unseen packet-loss rates at test time.

Our contributions are:
\begin{itemize}[leftmargin=*,nosep]
    \item We identify feature-level packet loss as a critical yet underexplored failure mode in collaborative perception and show that sparse receptions induce a distribution shift that existing methods do not explicitly address.
    \item We propose \methodname, a reconstruction framework that recovers missing BEV context from compact, packet-corrupted collaborator latents through prior-guided, ego-conditioned feature reconstruction.
    \item We show that training only under complete communication, combined with noise-conditioned prior refinement, can handle packet-loss-corrupted inputs at test time without packet-loss-specific training.
    \item We show that \methodname achieves the best overall mAP on both DAIR-V2X and V2XREAL under 90\% packet loss while using $64\times$ less bandwidth than dense BEV feature fusion baselines.
\end{itemize}

\FloatBarrier

\section{Related Work}

\subsection{Collaborative Perception}
Collaborative perception enables agents to share sensory information and overcome individual perceptual limitations. The prevailing paradigm is intermediate fusion, which exchanges BEV feature maps to balance communication efficiency and fusion effectiveness, avoiding the prohibitive bandwidth of raw-data exchange \cite{Cooper1,Cooper2} while surpassing late fusion. Early works used graph neural network (GNN)-based message passing \cite{v2vnet}; AttnFuse \cite{OPV2V} introduced attention-based inter-agent modeling, later scaled to global context via V2X-ViT \cite{v2x-vit}. DiscoNet \cite{disconet} explored knowledge distillation as a complementary enrichment strategy.

Robustness to real-world deployment uncertainties has been studied along several axes: pose errors \cite{localizationproblem, Coalign, ROCO, sparsealign, conformal_uncertainty}, communication latency \cite{latency1,latency2,Traf-align, CoBEVFlow}, and heterogeneous agent configurations \cite{CoBEVT, CoCa3D, v2vformer++, cocmt, bm2cp, HEAL, HM-VIT, stamp}. These efforts address localization imprecision, asynchrony, and cross-agent incompatibilities, which are orthogonal to the \emph{transmission reliability} problem we study.

Dense BEV feature fusion baselines also pose bandwidth challenges. Where2comm \cite{where2comm} learns to select salient regions, while instance-level methods further reduce bandwidth via object-centric representations \cite{TransIFF, quest, GSCOOP, ACCO}. CodeFilling \cite{CodeFilling} frames collaboration as compression-reconstruction, extended by QuantV2X \cite{quantv2x} with quantized representations. More recently, generative approaches such as DiffCP \cite{diffcp} and CoDiff \cite{codiff} utilize diffusion models to restore compact features, but assume reliable delivery and focus on reconstructing from designed compression artifacts. In contrast, our work addresses a distinct and complementary problem: reconstructing from \emph{stochastic packet loss} on already-compact latents, where the missingness pattern is unknown at training time and varies per transmission.

\subsection{Generative Reconstruction for Collaborative Perception}
When transmitted BEV features are aggressively compressed or spatially corrupted by packet loss, recovering complete, detection-ready features on the receiver side becomes the central challenge. Diffusion models offer a principled mechanism for this: from DDPM \cite{ddpm} through efficient samplers such as DDIM \cite{DDIM} and latent-space variants \cite{LDM}, they learn to reverse noise and synthesize structured signals from corrupted inputs. Conditioning mechanisms including ILVR \cite{ilvr}, ControlNet \cite{ControlNet}, and RePaint \cite{Repaint} extend this to reconstruction from partial or spatially incomplete observations, making them well-suited for feature recovery in V2X settings.

These properties have motivated applying generative reconstruction in collaborative perception. CodeFilling \cite{CodeFilling} and QuantV2X \cite{quantv2x} treat collaboration as compression-reconstruction using discrete codebooks. DiffCP \cite{diffcp} uses a diffusion model to reconstruct individual-agent features from ultra-compact representations before fusion; CoDiff \cite{codiff} applies diffusion to denoise fused features degraded by pose estimation errors; GenComm \cite{GenComm} employs a conditional diffusion model to generate features in the ego's semantic space, bridging domain gaps across heterogeneous agents. However, each of these methods targets a specific, pre-defined form of degradation and assumes reliable feature delivery. Our work instead targets stochastic packet loss on compact latents, recovering missing BEV content without assuming a specific loss pattern.

\section{Method}

\subsection{Problem Formulation}
\label{problem_formulation}
\textbf{Scenario.}
We consider a V2X collaborative perception system with $N$ agents $\mathcal{A} = \{A_0, A_1, \ldots, A_{N-1}\}$ communicating over a wireless network. Agent $A_0$ denotes the ego vehicle. Each agent $A_i$ operates in a local coordinate frame and observes the scene through onboard sensors, producing bird's-eye-view (BEV) features $X_i \in \mathbb{R}^{H \times W \times C}$. The spatial transformation from agent $A_i$ to agent $A_j$ is represented by an affine matrix $T_{i \to j} \in \mathbb{R}^{2 \times 3}$, which stacks a $2\times 2$ linear part (rotation/scale) and a $2\times 1$ translation vector.

\textbf{Communication Graph.}
Agents communicate over a directed graph $\mathcal{G} = (\mathcal{A}, \mathcal{P})$, where an edge $(i, j) \in \mathcal{P}$ exists if agent $A_i$ can transmit to agent $A_j$. Each agent $A_i$ transmits a message $S_i$ derived from its local BEV feature $X_i$, subject to a communication budget:
\begin{equation}
S_i = \mathcal{C}(X_i), \qquad \text{size}(S_i) \leq B,
\end{equation}
where $\mathcal{C}(\cdot)$ denotes a generic communication function and $B$ is the maximum bandwidth budget.

\textbf{Packet Loss Model.}
During transmission over edge $(i, j)$, we consider feature-level packet loss at the spatial-packet granularity. Let $M_{i \to j}$ denote the reception mask defined on the spatial layout of the transmitted message, where $M_{i \to j}(u, v) = 1$ indicates successful reception of the packet at spatial location $(u, v)$. We assume the following random packet corruption process:
\begin{equation}
M_{i \to j}(u, v) \sim \text{Bernoulli}(p), \quad \tilde{S}_{i \to j} = S_i \odot M_{i \to j},
\end{equation}
where $p \in [0, 1]$ is the retention rate and $\odot$ denotes element-wise multiplication across all channels. We refer to $\tilde{S}_{i \to j}$ as the received message. In the ego-centric setting, we use the shorthand $\tilde{S}_i \equiv \tilde{S}_{i \to 0}$ for $i \neq 0$, where the subscript $0$ denotes the ego agent.

\textbf{Optimization Objective.}
For each ego vehicle $A_0$, the goal is to perform collaborative 3D object detection from its own BEV feature and the messages received from other agents. The ego uses its own feature directly, while each collaborator feature is reconstructed in the collaborator coordinate frame from the received message, the ego feature, and the relative transform:
\begin{equation}
\hat{X}_0 = X_0, \qquad \hat{X}_i = \mathcal{R}(\tilde{S}_i, X_0, T_{0 \to i}; \bm{\theta}_r), \quad \forall i \neq 0,
\end{equation}
where $\bm{\theta}_r$ is the learnable parameter of the reconstruction model $\mathcal{R}$. Then, before fusion, each reconstructed collaborator feature is warped back to the ego coordinate frame, denoted by $\bar{X}_i = \mathcal{W}(\hat{X}_i, T_{i \to 0})$ for $i \neq 0$ and $\bar{X}_0=X_0$. These ego-aligned features are fused for ego-centric detection:
\begin{equation}
\hat{Y} = \mathcal{D}\left( \mathcal{F}\left( \{\bar{X}_i\}_{i=0}^{N-1}; \bm{\theta}_f \right); \bm{\theta}_d \right).
\end{equation}
Here $\bm{\theta}_f$ and $\bm{\theta}_d$ denote the learnable parameters of the multi-agent fusion module $\mathcal{F}$ and the detection head $\mathcal{D}$, respectively.
The training objective minimizes the detection loss
\begin{equation}
\mathcal{L}_{\text{det}}(\hat{Y}, Y),
\end{equation}
where $Y$ denotes ground-truth object annotations. We evaluate the robustness of the above collaborative perception process under varying retention rates $p$ and random masks $M$ to reflect stochastic packet loss during transmission. 

\subsection{Framework Overview}
\begin{figure}[htbp]
  \centering
  \includegraphics[width=\linewidth]{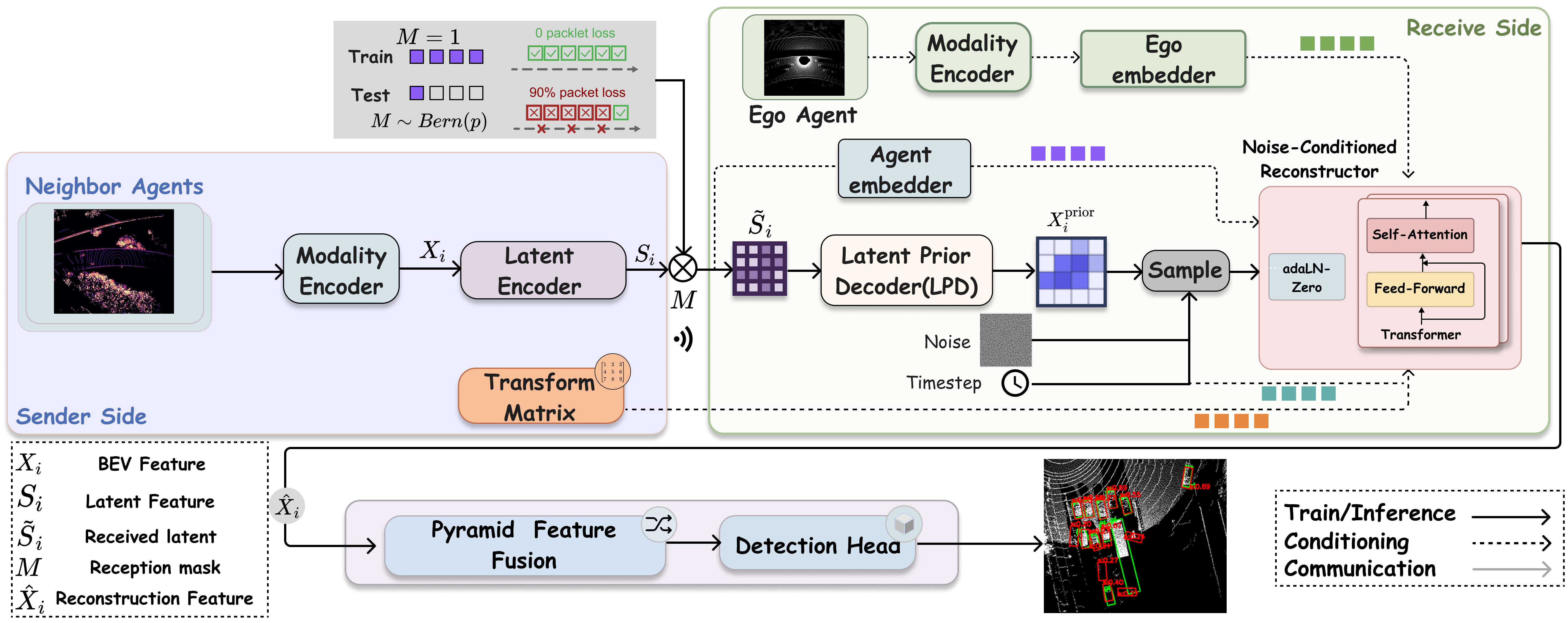}
  \caption{Framework of \methodname. Each agent encodes BEV features into a compact latent for transmission. The receiver reconstructs the agent's BEV via the Latent Prior Decoder and a noise-conditioned reconstructor, then fuses with its own features for 3D detection.}
  \label{Framework}
\end{figure}
The received message is simultaneously (i)~compact after compression and (ii)~spatially incomplete after packet loss, making direct BEV regression from corrupted compact observations ill-conditioned. LR-V2X resolves this tension through a coarse-to-fine recovery pipeline. First, latent encoding reduces the transmitted BEV feature to a compact spatial carrier, establishing the low-bandwidth operating point. Second, the Latent Prior Decoder (LPD) converts the corrupted received latent into a dense but coarse BEV prior by propagating the remaining spatial evidence into neighboring missing regions. Third, a noise-conditioned reconstructor refines this prior using both the received latent and ego-side context. 

We instantiate the communication function as a learned latent encoder:
\begin{equation}
S_i = \mathcal{E}(X_i; \bm{\theta}_e) \in \mathbb{R}^{h \times w \times C'},
\end{equation}
where $\mathcal{E}(\cdot; \bm{\theta}_e)$ maps the collaborator BEV feature to a compact spatial latent with $h \ll H$, $w \ll W$, and $C' \leq C$. Under packet loss, the receiver observes $\tilde{S}_i = S_i \odot M_{i \to 0}$, a spatially incomplete version of this latent. The following subsections detail how the LPD converts $\tilde{S}_i$ into a dense prior and how the noise-conditioned reconstructor refines it into $\hat{X}_i$.

\subsection{Latent Prior Decoder}

The main purpose of the LPD is to make reconstruction well-posed before the transformer denoiser is applied. Packet loss turns the compact latent into a spatially incomplete observation whose missing locations provide no direct evidence, while a transformer operating directly on this corrupted compact input would have to infer both global structure and local completion from a weak initialization. The LPD therefore serves as a deterministic prior generator that converts the received latent $\tilde{S}_i$ into a spatially complete but coarse BEV map $X_i^{\text{prior}}$ and anchors the following reconstruction stage.

\begin{figure}[htbp]
  \centering
  \includegraphics[width=\linewidth]{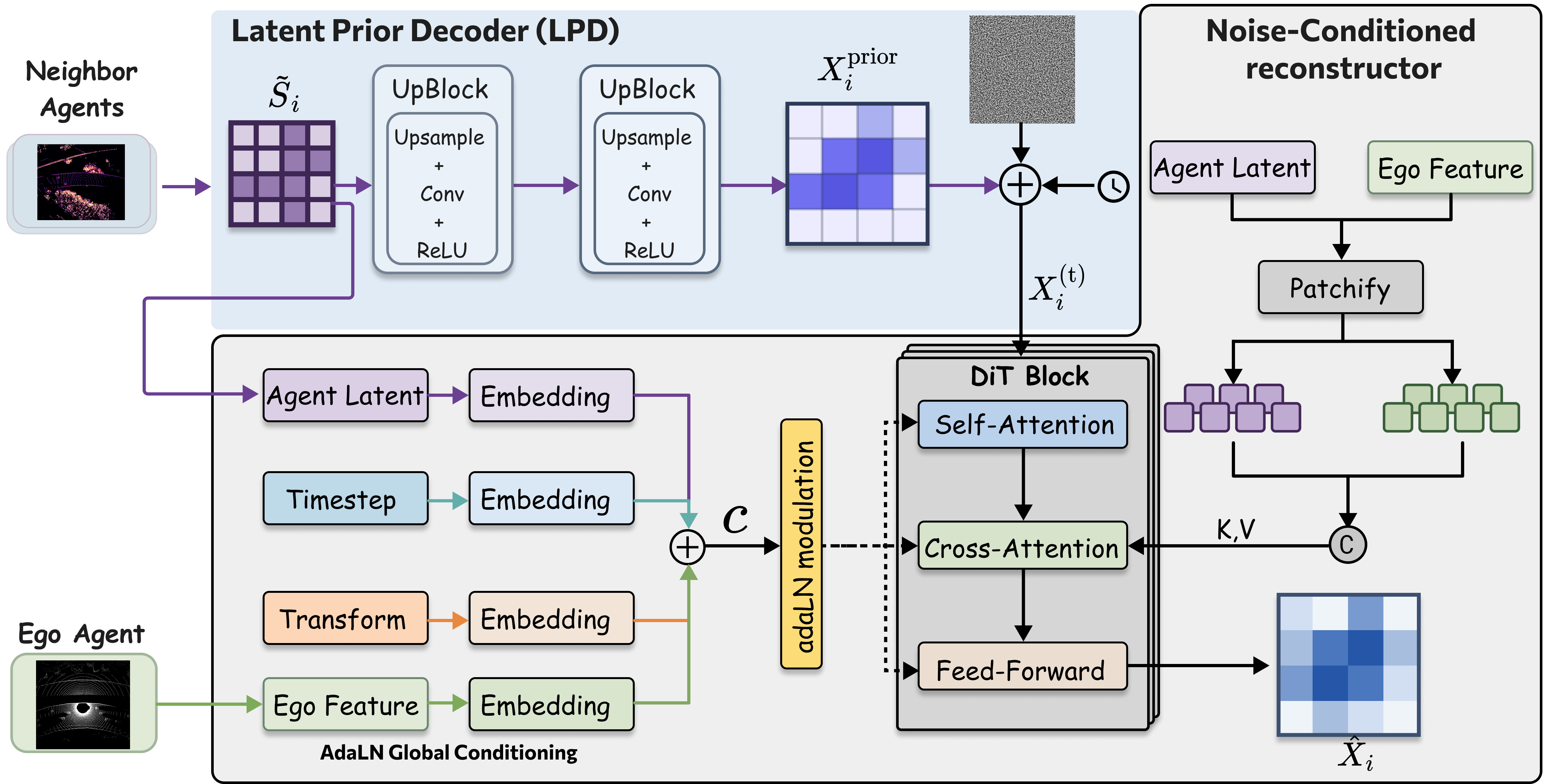}
  \caption{Architecture of the Latent Prior Decoder (LPD) and the DiT denoiser used for conditional feature reconstruction.}
  \label{block_diagram}
\end{figure}

Concretely, the LPD maps $\tilde{S}_i$ to $X_i^{\text{prior}}$ via four transposed convolutional layers \cite{FCN} with overlapping receptive fields, upsampling from latent resolution $(h, w)$ to BEV resolution $(H, W)$ (Fig.~\ref{block_diagram}):
\begin{equation}
X_i^{\text{prior}} = \text{LPD}(\tilde{S}_i).
\end{equation}
The overlapping kernels spread activations from received locations into neighboring missing regions, converting a spatially incomplete latent into a spatially complete prior BEV map~\cite{depth-complete,sparse2dense}. Although the LPD is trained on complete latents ($\tilde{S}_i = S_i$), the overlapping transposed convolutions still aggregate the available neighboring evidence when parts of the latent are missing, yielding a usable prior BEV map across arbitrary packet-loss rates.

\subsection{Noise-Conditioned Feature Reconstruction}

The noise-conditioned reconstructor is designed to refine an imperfect prior rather than generate BEV features from scratch. Its key design choice is to perturb the LPD prior itself during training. Unlike standard diffusion models, which add noise to the clean target and then learn an iterative generation process, our reconstruction process starts from the LPD prior because inference also starts from a packet-loss-degraded prior. This makes the learned denoising task match the actual test-time recovery problem and supports single-step feature reconstruction.

Given $\tilde{S}_i$, the LPD first produces the prior BEV map $X_i^{\text{prior}}$. Gaussian noise is then injected at a randomly sampled timestep $t$, yielding the noisy prior $X_i^{(t)}$:
\begin{equation}
X_i^{(t)} = \sqrt{\bar{\alpha}_t}\, X_i^{\text{prior}} + \sqrt{1 - \bar{\alpha}_t}\, \epsilon,
\end{equation}
where $\epsilon \sim \mathcal{N}(0, I)$ and $\bar{\alpha}_t$ is the cumulative noise schedule~\cite{ddpm}. A conditional denoiser $f_{\bm{\theta}_r}$ then recovers the clean BEV feature $\hat{X}_i$ from the noisy prior, conditioned on the received latent $\tilde{S}_i$, the per-collaborator aligned ego BEV $X_0^{\text{aligned}} = \mathcal{W}(X_0, T_{0 \to i})$ (warped from $X_0$ to agent $i$'s coordinate frame via bilinear interpolation), and the spatial transform $T_{0 \to i}$:
\begin{equation}
\hat{X}_i = f_{\bm{\theta}_r}(X_i^{(t)}, t, \tilde{S}_i, X_0^{\text{aligned}}, T_{0 \to i}).
\end{equation}
We train with an $x_0$-prediction objective:
\begin{equation}
  \label{L_diff}
\mathcal{L}_{\text{diff}} = \mathbb{E}_{t, \epsilon}\left[ \| f_{\bm{\theta}_r}(X_i^{(t)}, t, \tilde{S}_i, X_0^{\text{aligned}}, T_{0 \to i}) - X_i \|^2 \right].
\end{equation}

\subsubsection{Noise-Conditioned Reconstructor Architecture}

We adopt DiT \cite{DiT} as the denoiser backbone. The received latent $\tilde{S}_i$ constrains reconstruction in two complementary ways: it provides a global conditioning signal that steers the denoising trajectory toward the transmitted collaborator content, and it also serves as spatial evidence during cross-attention. The aligned ego BEV $X_0^{\text{aligned}}$ is introduced as additional cross-attention context, allowing the model to combine the transmitted message with ego-side scene structure when recovering missing regions. For layer $\ell$, the DiT denoiser block performs (Fig.~\ref{block_diagram})
\begin{equation}
  \begin{gathered}
  h_\ell = \operatorname{SelfAttn}(\operatorname{AdaLN}(x_\ell, c)) + x_\ell \\
  h'_\ell = \operatorname{CrossAttn}(\operatorname{AdaLN}(h_\ell, c),\, [\tilde{S}_i;\, X_0^{\text{aligned}}]) + h_\ell \\
  x_{\ell+1} = \operatorname{FFN}(\operatorname{AdaLN}(h'_\ell, c)) + h'_\ell.
  \end{gathered}
  \end{equation}
The global conditioning vector is
\begin{equation}
  c = \phi_t(t) + \phi_S(\operatorname{pool}(\tilde{S}_i)) + \phi_X(\operatorname{pool}(X_0^{\text{aligned}})) + \phi_T(T_{0 \to i}),
\end{equation}
where $\phi_t, \phi_S, \phi_X, \phi_T$ are learned MLP projections, and $\operatorname{AdaLN}(x, c)$ applies layer normalization with per-channel scale and shift linearly projected from $c$. Cross-attention allows the model to query both transmitted evidence and ego context when reconstructing missing regions. This combination lets the denoiser preserve global consistency with the received message while restoring local object structure in corrupted regions.

\newcommand{\performanceTable}{%
\begin{table*}[htbp]
  \caption{Performance comparison under \textbf{Complete Communication (0\% Packet Loss)} reference conditions and the main \textbf{90\% Packet Loss} robustness setting.}
  \label{tab:performance}
  \begin{subtable}{\textwidth}
    \centering
    \caption{V2XREAL dataset (per-class AP and mAP).}
    \label{tab:overall_comparison}
    \resizebox{\textwidth}{!}{
    \begin{tabular}{l|ccc|ccc|ccc|ccc|cc}
        \toprule
        \multirow{2}{*}{\textbf{Models}} & \multicolumn{3}{c|}{\textbf{AP\_vehicle}} & \multicolumn{3}{c|}{\textbf{AP\_pedestrian}} & \multicolumn{3}{c|}{\textbf{AP\_truck}} & \multicolumn{3}{c|}{\textbf{mAP (\%)}} & \multirow{2}{*}{\textbf{Comm}} & \multirow{2}{*}{\textbf{FPS}} \\
        \cmidrule(lr){2-4} \cmidrule(lr){5-7} \cmidrule(lr){8-10} \cmidrule(lr){11-13}
        & 0.3 & 0.5 & 0.7 & 0.3 & 0.5 & 0.7 & 0.3 & 0.5 & 0.7 & 0.3 & 0.5 & 0.7 & & \\
        \midrule
        \rowcolor{grouphdr}
        \multicolumn{15}{c}{\textbf{No Packet Loss}} \\
        \midrule
        No Fusion & 32.27 & 17.20 & 1.31 & 36.39 & 30.91 & 10.79 & 55.11 & 52.76 & 37.14 & 41.26 & 33.63 & 16.41 & 0 & 2.47 \\
        Early Fusion & 68.17 & 64.73 & 40.03 & 33.53 & 17.14 & 0.87 & 27.22 & 25.57 & 14.18 & 42.97 & 35.81 & 18.36 & $\sim$2.4MB & 1.03 \\
        Late Fusion & 32.54 & 16.23 & 1.17 & 34.12 & 30.13 & 13.70 & 61.55 & 58.96 & 38.99 & 42.73 & 35.10 & 17.96 & $\leq$5.4KB & 1.21 \\
        \rowcolor{oursrow}
        \methodname & 76.70 & 73.85 & 49.58 & 35.98 & 17.40 & 1.12 & 48.38 & 43.60 & 32.64 & 53.69 & 44.95 & 27.78 & 128KB & 1.54 \\
        \midrule
        \rowcolor{grouphdr}
        \multicolumn{15}{c}{\textbf{90\% Packet Loss}} \\
        \midrule
        Attn & 59.35 & 55.93 & 39.25 & 29.30 & 15.20 & 1.08 & 39.00 & 35.50 & 20.66 & 42.55 & 35.54 & 20.33 & 8MB & 1.24 \\
        F-Cooper & 57.94 & 54.22 & 38.00 & 31.92 & 16.31 & 1.08 & 37.60 & 32.40 & 12.60 & 42.49 & 34.30 & 17.21 & 8MB & 0.82 \\
        CoBEVT & 51.54 & 46.45 & 30.60 & 25.25 & 11.35 & 0.50 & 39.99 & 34.98 & 15.82 & 38.92 & 30.93 & 15.64 & 8MB & 0.82 \\
        V2X-VIT & 49.37 & 46.18 & 31.40 & \textbf{34.23} & \textbf{17.77} & \textbf{1.38} & 38.60 & 32.90 & 24.00 & 40.73 & 32.29 & 18.93 & 8MB & 0.70 \\
        CodeFilling & 62.14 & 57.95 & 39.36 & 16.41 & 12.92 & 0.60 & \textbf{43.07} & 39.19 & \textbf{29.52} & 40.54 & 36.69 & 23.16 & 128KB & \textbf{1.68} \\
        HEAL & 62.31 & 60.85 & 39.29 & 26.53 & 12.96 & 0.99 & 40.36 & 39.09 & 27.10 & 43.07 & 37.64 & 21.79 & 8MB & 1.61 \\
        GenComm$^\dagger$ & 63.11 & 56.60 & 30.38 & 12.99 & 3.83 & 0.15 & 33.33 & 29.97 & 17.48 & 36.48 & 30.14 & 16.00 & 128KB & 0.37 \\
        \rowcolor{oursrow}
        \methodname & \textbf{64.60} & \textbf{63.59} & \textbf{46.31} & 28.55 & 12.23 & 1.09 & 41.48 & \textbf{39.98} & 25.12 & \textbf{44.87} & \textbf{38.60} & \textbf{24.18} & 128KB & 1.54 \\
        \bottomrule
    \end{tabular}
    }
  \end{subtable}
  \vspace{0.6em}
  \begin{subtable}{\textwidth}
    \centering
    \caption{DAIR-V2X dataset (per-range AP and mAP).}
    \label{tab:dair_v2x_comparison}
    \resizebox{\textwidth}{!}{
    \begin{tabular}{l|ccc|ccc|ccc|ccc|cc}
        \toprule
        \multirow{2}{*}{\textbf{Models}} & \multicolumn{3}{c|}{\textbf{Short}} & \multicolumn{3}{c|}{\textbf{Middle}} & \multicolumn{3}{c|}{\textbf{Long}} & \multicolumn{3}{c|}{\textbf{mAP (\%)}} & \multirow{2}{*}{\textbf{Comm}} & \multirow{2}{*}{\textbf{FPS}} \\
        \cmidrule(lr){2-4} \cmidrule(lr){5-7} \cmidrule(lr){8-10} \cmidrule(lr){11-13}
        & 0.3 & 0.5 & 0.7 & 0.3 & 0.5 & 0.7 & 0.3 & 0.5 & 0.7 & 0.3 & 0.5 & 0.7 & & \\
        \midrule
        \rowcolor{grouphdr}
        \multicolumn{15}{c}{\textbf{No Packet Loss}} \\
        \midrule
        No Fusion & 79.25 & 77.24 & 70.12 & 75.70 & 72.49 & 62.20 & 54.78 & 51.06 & 37.94 & 62.51 & 59.33 & 49.67 & 0 & 8.70 \\
        Early Fusion & 84.47 & 81.40 & 68.88 & 78.60 & 72.88 & 56.54 & 59.79 & 54.55 & 36.13 & 73.52 & 68.61 & 52.37 & $\sim$1.0MB & 14.95 \\
        Late Fusion & 79.25 & 77.24 & 70.12 & 75.70 & 72.49 & 62.20 & 54.78 & 51.06 & 37.94 & 69.51 & 66.33 & 55.67 & $\leq$5.4KB & 7.75 \\
        \rowcolor{oursrow}
        \methodname & 88.48 & 85.84 & 72.00 & 83.56 & 78.38 & 58.51 & 71.56 & 66.04 & 46.16 & 81.03 & 76.16 & 57.74 & 128KB & 6.04 \\
        \midrule
        \rowcolor{grouphdr}
        \multicolumn{15}{c}{\textbf{90\% Packet Loss}} \\
        \midrule
        Attn & 80.53 & 79.02 & 66.72 & \textbf{77.47} & \textbf{71.70} & 56.83 & \textbf{57.23} & 51.41 & \textbf{38.45} & 66.87 & 66.20 & 52.56 & 8MB & 6.67 \\
        F-Cooper & 80.43 & 77.14 & 67.13 & 76.96 & 71.62 & 57.04 & 57.15 & \textbf{51.65} & 38.44 & 67.89 & 65.90 & 52.87 & 8MB & 5.39 \\
        CoBEVT & 79.98 & 76.17 & 61.25 & 75.15 & 69.79 & 51.91 & 53.46 & 47.68 & 33.25 & 68.83 & 63.55 & 44.47 & 8MB & 4.92 \\
        V2X-VIT & 80.01 & 77.06 & 66.55 & 75.34 & 71.04 & 56.20 & 53.23 & 48.04 & 34.61 & 68.78 & 64.33 & 50.97 & 8MB & 4.70 \\
        CodeFilling & 76.92 & 75.24 & 66.22 & 69.68 & 67.03 & 55.20 & 47.43 & 44.80 & 35.29 & 63.38 & 60.93 & 50.62 & 128KB & \textbf{9.17} \\
        HEAL & 76.88 & 75.26 & 65.77 & 70.48 & 67.64 & 54.60 & 49.39 & 46.20 & 35.69 & 64.43 & 61.68 & 50.50 & 8MB & 7.24 \\
        GenComm$^\dagger$ & 67.89 & 62.56 & 37.75 & 49.68 & 44.51 & 22.82 & 32.00 & 27.81 & 13.20 & 48.03 & 43.16 & 23.27 & 128KB & 0.49 \\
        \rowcolor{oursrow}
        \methodname & \textbf{82.59} & \textbf{80.93} & \textbf{69.98} & 75.88 & 70.03 & \textbf{57.93} & 55.25 & 49.86 & 36.96 & \textbf{69.35} & \textbf{66.90} & \textbf{53.37} & 128KB & 6.04 \\
        \bottomrule
    \end{tabular}
    }
  \end{subtable}
  \vspace{0.3em}
  {\raggedright\footnotesize $^\dagger$ Re-implementation of GenComm's DDIM reconstruction module under our backbone and training protocol.\par}
  {\raggedright\footnotesize Communication cost is reported per transmitted collaborator frame before packet loss. Raw-point early fusion uses float32 XYZI payloads estimated from validation files (V2XREAL: $\sim$2.4\,MB; DAIR-V2X: $\sim$1.0\,MB). Dense BEV and compact latent messages use float32 tensor sizes. Late fusion reports a compact 7-DoF box, score, and class payload with the post-NMS cap of 150 boxes, giving $\leq$5.4\,KB.\par}
\end{table*}%
}

\subsubsection{Training and Inference}

Training follows a stage-wise protocol. We first train the PointPillar encoder, pyramid fusion backbone, and detection heads under complete communication. We then freeze the sensor encoder/backbone and train the latent encoder $\mathcal{E}$, LPD, and DiT reconstructor by minimizing $\mathcal{L}_{\text{diff}}$ (Eq.~\ref{L_diff}) under complete communication ($\tilde{S}_i = S_i$), with $t$ sampled uniformly from $\{1, \ldots, T\}$ at each step. Finally, we fine-tune the reconstruction module together with the fusion and detection heads while keeping the sensor encoder/backbone frozen. This exposes the denoiser to priors of varying quality without requiring packet-loss-corrupted messages during training.
At inference, given $\tilde{S}_i$ from agent $i$ ($i \neq 0$), we compute $X_i^{\text{prior}} = \text{LPD}(\tilde{S}_i)$, draw Gaussian noise, perturb the prior to a fixed noise level $t_{\text{inf}}$ (a hyperparameter), and recover $\hat{X}_i$ in a single forward pass of $f_{\bm{\theta}_r}$. The reconstructed collaborator feature is then aligned back to the ego frame inside the fusion module. For end-to-end optimization with the detection task, the total loss combines reconstruction and detection objectives
\begin{equation}
\mathcal{L}_{\text{total}} = \lambda_\text{det}\mathcal{L}_{\text{det}}(\mathcal{D}(\mathcal{F}(\{\bar{X}_i\})), Y) + \lambda_\text{diff} \mathcal{L}_{\text{diff}},
\end{equation}
where $\lambda_\text{det}, \lambda_\text{diff}$ are the balancing weights.

\FloatBarrier

\section{Experiments}

\subsection{Setup}
\label{sec:setup}
We evaluate on two real-world V2X datasets: DAIR-V2X \cite{dair-v2x} and V2XREAL \cite{v2xreal}. All methods use a PointPillars \cite{PointPillars} backbone and are trained under complete communication (no packet loss). At test time, we simulate spatial-level packet loss by independently dropping each spatial packet with probability $1 - p$, where $p$ is the retention rate. This i.i.d. Bernoulli model provides a controlled testbed for random packet erasures; we additionally evaluate correlated burst loss in Appendix~\ref{app:burst_loss}. Packet-loss masks are applied only to transmitted collaborator messages, while the ego feature remains available locally to the receiver. We use fixed random seeds for mask generation and inference noise, so all methods are compared under the same stochastic protocol. We measure dataset-specific AP breakdowns and overall mAP at IoU thresholds 0.3, 0.5, and 0.7, along with per transmitted collaborator frame communication cost and inference speed (frames per second, FPS). Full implementation details, hyperparameters, and training procedures are provided in Appendix~\ref{app:setup}.

\subsection{Performance}
\label{sec:performance}

\performanceTable

Table~\ref{tab:performance} reports dataset-specific AP breakdowns and overall mAP at IoU 0.3, 0.5, and 0.7, along with communication cost and inference speed. On V2XREAL (Table~\ref{tab:overall_comparison}), \methodname achieves the highest overall mAP across all three IoU thresholds under 90\% packet loss (44.87/38.60/24.18), while transmitting only a 128\,KB collaborator message ($64\times$ less than dense BEV feature fusion baselines). Compared to CodeFilling at the same 128\,KB budget, \methodname improves mAP@0.5 by 1.91 and mAP@0.7 by 1.02 points, and it also outperforms the protocol-matched GenComm$^\dagger$ re-implementation. On DAIR-V2X (Table~\ref{tab:dair_v2x_comparison}), \methodname likewise achieves the highest overall mAP across all three IoU thresholds (69.35/66.90/53.37), surpassing all dense BEV feature fusion baselines that require $64\times$ more communication. Under complete communication, \methodname also attains the highest overall mAP among the no-packet-loss references reported here on both datasets.

\begin{figure}[htbp]
  \centering
  \includegraphics[width=\linewidth]{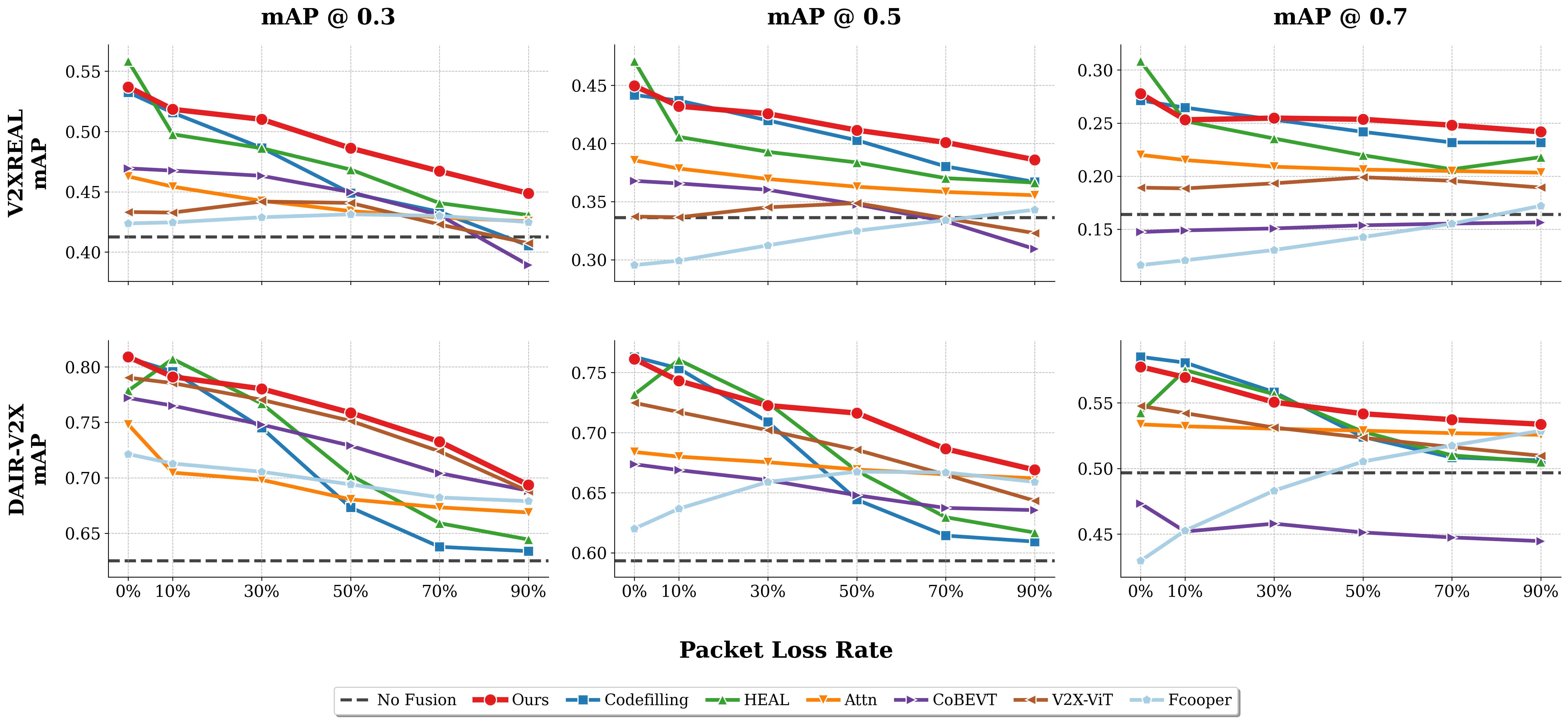}
  \caption{Performance under varying packet loss rates on DAIR-V2X and V2XREAL. Across the sweep, \methodname remains consistently competitive as communication quality decreases and reaches the highest overall mAP at the 90\% packet-loss endpoint.}
  \label{pk_loss_compare_fig}
\end{figure}

Fig.~\ref{pk_loss_compare_fig} shows performance across packet loss rates from 0\% to 90\%. \methodname is not the top method at every point in the sweep, but it remains consistently competitive across IoU thresholds and datasets, ending with the highest overall mAP under 90\% packet loss.

\textbf{Bandwidth--robustness asymmetry.}
Dense BEV feature fusion baselines transmit 8\,MB per collaborator message ($64\times$ our budget), so at 90\% loss they still retain 819.2\,KB of spatial information versus 12.8\,KB for a 128\,KB latent. This communication redundancy provides an inherent robustness advantage that is not attributable solely to fusion design (see Appendix~\ref{app:bandwidth_robustness}). Appendix Table~\ref{tab:budget_equalized_dense} controls for this directly: after matching dense BEV feature fusion baselines to 128\,KB, all five decrease in mAP@0.5, while \methodname at the same budget achieves the best overall mAP (69.35/66.90/53.37). This supports the interpretation that explicit reconstruction, not only raw communication volume, is responsible for the low-bandwidth robustness.

Two complementary robustness controls further support this interpretation. Appendix Table~\ref{tab:burst_loss} replaces i.i.d. erasures with correlated burst loss; at the same 128\,KB budget, \methodname still outperforms CodeFilling by +2.47/+1.59/+0.45 mAP at IoU 0.3/0.5/0.7. Together with the budget-matched dense BEV feature fusion results, this shows that the low-bandwidth gain is not limited to independent random drops or raw communication volume.

\subsection{Ablation Study}

\label{sec:ablation}

\textbf{Conditioning inputs.} Appendix Table~\ref{tab:ablation} ablates the reconstructor conditions while keeping the latent-derived prior fixed. Ego-BEV is the decisive condition: adding it raises mAP@0.3 by 11.62 points, whereas the transformation matrix adds 0.79 points on top of ego-BEV and the direct latent branch alone changes mAP@0.3 by only 0.02 points. This is consistent with V2X geometry: the ego BEV often overlaps the collaborator field of view, so after alignment it provides local anchors that help the reconstructor disambiguate missing or corrupted collaborator features.

\textbf{Latent Prior Decoder.} Appendix Fig.~\ref{ablation_lpd_fig} compares the full model with two controlled variants: prior-only uses the LPD output directly for fusion and skips the denoising reconstructor, while zero-prior keeps the denoising reconstructor but replaces the LPD initialization with an all-zero BEV prior before the same noise-conditioned inference step. The full model remains consistently stronger, indicating that LPD initialization and denoising refinement are complementary rather than redundant.

\textbf{Downsampling ratio.} Appendix Table~\ref{tab:downsample} shows that moving from $\times$2 (2\,MB) to $\times$8 (128\,KB) reduces mAP@0.3 by only 1.52 points while reducing bandwidth $16\times$; at $\times$16 (32\,KB), mAP@0.3 drops to 79.58. We therefore use $\times$8 as the default accuracy--bandwidth tradeoff.

\textbf{Inference timestep and noise-conditioned training.} Appendix Table~\ref{tab:nonoise_compare} shows that noise conditioning improves over a protocol-matched no-noise reconstructor by +7.16/+7.22/+4.23 mAP. This no-noise variant still uses the LPD prior and a learned reconstructor, but feeds the prior directly at $t=0$ without the forward-noise perturbation. Appendix Fig.~\ref{fig:tinf_sweep} further shows broadly stable performance with $t_{\text{inf}} = 700$ selected from the sweep.

Taken together, these ablations support a clear division of labor. The compact latent preserves spatially aligned collaborator evidence, the LPD turns surviving packets into a complete but coarse prior, and the noise-conditioned reconstructor refines this prior with ego context instead of treating it as a direct decoding output. This explains why the default configuration remains effective under high packet loss without increasing the 128\,KB message budget.

\section{Conclusion}

This work studies V2X collaborative perception under the joint constraint of limited bandwidth and stochastic packet loss. We introduced \methodname, a reconstruction-first framework that transmits compact collaborator latents and recovers missing BEV context through an LPD prior and an ego-conditioned, noise-conditioned reconstructor. The experiments show that explicit feature recovery from corrupted compact messages provides stronger low-bandwidth robustness than relying on dense communication redundancy or direct compact-code fusion alone. A central implication is that packet-loss robustness can be learned by modeling how to refine imperfect latent priors, even when training uses complete communication rather than enumerating many lossy conditions. Future work can extend this reconstruction-first view to richer wireless channel models.

{
    \small
    \bibliographystyle{plainnat}
    \bibliography{references}
}

\clearpage
\appendix

\section{Experimental Protocol and Implementation Details}
\label{app:setup}

This appendix follows the reading path of the main paper. Appendix~\ref{app:setup} documents the datasets, packet-loss protocol, and implementation details used in Sec.~\ref{sec:setup}. Appendix~\ref{app:robustness} extends the robustness discussion in Sec.~\ref{sec:performance}. Appendix~\ref{app:additional_experiments} expands the ablations summarized in Sec.~\ref{sec:ablation}. Appendix~\ref{app:design_clarifications} clarifies two method choices from Sec.~\ref{problem_formulation}, and Appendix~\ref{app:limitations} states the experimental scope and future extensions.

\subsection{Datasets and Metrics}
We evaluate \methodname on two real-world collaborative perception benchmarks for LiDAR-based cooperative 3D object detection: DAIR-V2X \cite{dair-v2x} and V2XREAL \cite{v2xreal}. DAIR-V2X represents the vehicle-to-infrastructure setting, while V2XREAL provides real-world vehicle-side collaborative driving scenes and is evaluated here in the LiDAR-only V2V setting used by OpenCOOD-style baselines. We follow the standard benchmark training and evaluation protocol for each dataset and preserve the native metric breakdown of each benchmark in the main tables: DAIR-V2X is reported with range-based AP, and V2XREAL is reported with class-based AP. Unless otherwise stated, all compared methods use the same PointPillar backbone so that the experiments isolate the effect of communication and reconstruction under packet loss.

For DAIR-V2X, Short, Middle, and Long denote diagnostic AP values computed on objects within the corresponding distance ranges, while the overall mAP columns are computed from all evaluated detections across the full range and are not the arithmetic mean of the three range bins. For V2XREAL, the overall mAP is the mean of the unrounded AP values over the vehicle, pedestrian, and truck super-classes.

\subsection{Packet-Loss Evaluation Protocol}
Unless otherwise stated, all methods are trained under complete communication and evaluated under the same packet-loss simulation protocol. During testing, each spatial packet on the transmitted communication tensor is independently dropped with probability $1 - p$, where $p$ is the retention rate. The main robustness setting uses $p = 0.1$ (90\% packet loss), and Fig.~\ref{pk_loss_compare_fig} sweeps the corresponding packet-loss rate from 0\% to 90\%. The ego feature is not transmitted and is always available to the ego-side fusion module. Evaluation uses fixed random seeds for mask generation and inference noise so that methods are compared under the same stochastic protocol.

\subsection{Implementation Details}
Our experiments are implemented in the OpenCOOD \cite{OPV2V} framework. We use PointPillar \cite{PointPillars} as the encoder with a voxel size of ($0.4$m, $0.4$m) and a maximum of 32 points per pillar over the LiDAR range $[-102.4, -51.2, -3.5, 102.4, 51.2, 1.5]$m. After PointPillars and backbone downsampling, the BEV feature tensor used by the latent encoder has spatial resolution $128 \times 256$ with 64 channels. All methods use the same evaluation protocol with fixed packet-loss masks and inference seeds. Configuration files and scripts for the main training and evaluation settings will be released with the code.

\subsubsection{Training Pipeline}
We adopt a three-stage training pipeline:

\textbf{Stage 1: Baseline Detection.}
We train the PointPillar encoder, pyramid fusion backbone (ResNeXt with 3 scales: $[64, 128, 256]$ channels, upsample strides $[1, 2, 4]$), and detection head for 40 epochs with batch size 8. The optimizer is Adam with learning rate $2 \times 10^{-3}$, L2 regularization implemented via weight decay $10^{-4}$, and a multistep LR schedule (decay $\times 0.1$ at epochs 15 and 30). The detection loss combines Sigmoid Focal Loss ($\alpha=0.25$, $\gamma=2.0$) for classification, Weighted Smooth L1 Loss ($\sigma=3.0$) for regression, and pyramid losses at relative downsamples $[1, 2, 4]$ with weights $[0.4, 0.2, 0.1]$.

\textbf{Stage 2: Reconstructor Training.}
We freeze the PointPillar sensor encoder, pyramid fusion backbone, and detection heads, and train the compact latent encoder together with the LPD and DiT denoiser for 100 epochs with batch size 12. The optimizer is Adam with learning rate $2 \times 10^{-4}$, L2 regularization implemented via weight decay $10^{-4}$, and a multistep LR schedule (decay $\times 0.1$ at epochs 50 and 80). Only the $x_0$-prediction MSE loss is used in this stage.

\textbf{Stage 3: End-to-End Fine-tuning.}
We jointly fine-tune the latent encoder, LPD, DiT denoiser, pyramid fusion module, and detection heads while keeping the PointPillar sensor encoder/backbone frozen. This stage runs for 20 epochs with batch size 8. The total loss is $\mathcal{L}_{\text{total}} = \mathcal{L}_{\text{det}} + 0.1 \cdot \mathcal{L}_{\text{diff}}$, using Adam with learning rate $2 \times 10^{-4}$ and the same L2 regularization setting.

\subsubsection{Architecture Details}

\textbf{Latent Encoder.}
The encoder compresses BEV features from $128 \times 256 \times 64$ to a target spatial resolution of $16 \times 32$ with 64 latent channels (default $\times$8 downsampling), yielding a 128\,KB message per transmitted collaborator frame.

\textbf{Communication Accounting.}
For compact latent communication, the default payload contains $16 \times 32 \times 64$ float32 values. Dense BEV feature fusion transmits the corresponding $128 \times 256 \times 64$ float32 tensor, giving 8\,MB. Early fusion sends raw point clouds, so we estimate its payload from validation files under float32 XYZI storage: about 1.0\,MB on DAIR-V2X and about 2.4\,MB on V2XREAL. Late fusion sends post-NMS detections; using the implementation cap of 150 boxes and a compact 7-DoF box, score, and class payload gives an upper bound of 5.4\,KB.

\textbf{Latent Prior Decoder (LPD).}
The LPD consists of four transposed convolutional layers that upsample from latent resolution $(16, 32)$ back to BEV resolution $(128, 256)$. Overlapping kernels enable spatial propagation from received locations into missing regions.

\textbf{DiT Denoiser.}
We use the DiT-B configuration: 12 transformer layers, hidden dimension 768, 12 attention heads, MLP ratio 4.0, and patch size 4. Cross-attention layers attend to concatenated tokens from the received latent and aligned ego BEV. Gradient checkpointing is enabled to reduce memory usage.

\subsubsection{Noise Schedule and Inference Configuration}
We use a cosine noise schedule with 1000 timesteps ($\beta_{\text{start}} = 10^{-4}$, $\beta_{\text{end}} = 0.02$). At inference, we sample Gaussian noise for the LPD prior, perturb it to a fixed inference timestep $t_{\text{inf}} = 700$, and perform a single DDIM-style \cite{DDIM} denoising forward pass from the noise-augmented prior to the reconstructed BEV feature. We keep this timestep fixed for all datasets and packet-loss rates, so the reported results do not rely on condition-specific tuning.

\section{Robustness Analyses Beyond the Main Sweep}
\label{app:robustness}

This section extends the performance analysis in Sec.~\ref{sec:performance}. We first evaluate correlated burst loss to check whether the packet-loss trend is tied to the i.i.d.\ erasure model, and then use bandwidth-matched comparisons to separate reconstruction quality from the amount of information retained after transmission.

\subsection{Correlated Burst Loss}
\label{app:burst_loss}

\begin{figure}[!t]
  \centering
  \includegraphics[width=0.82\linewidth]{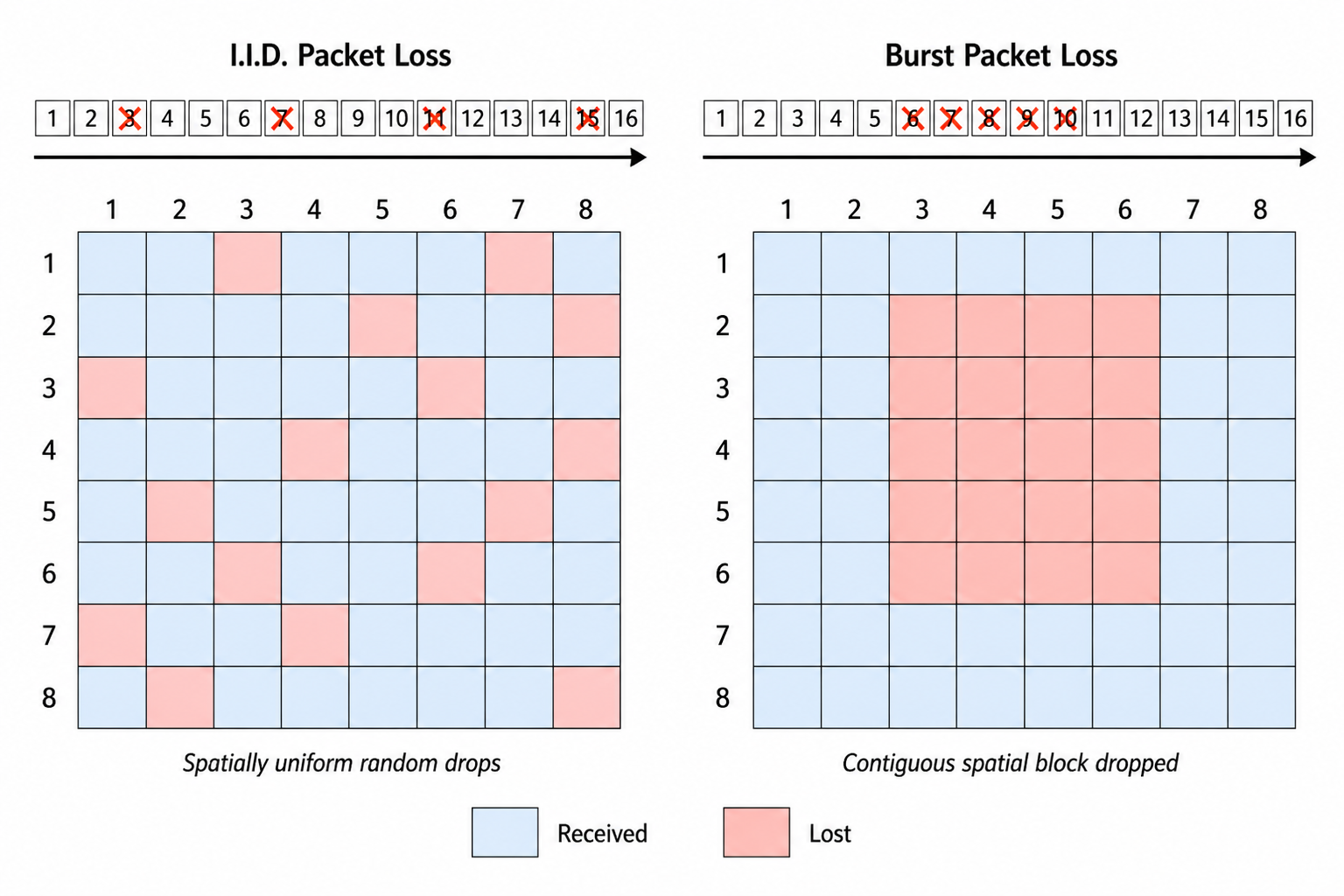}
  \caption{Illustration of i.i.d.\ packet loss (left) vs.\ burst packet loss (right) on an $8\times8$ BEV spatial grid. Blue cells are received; red cells are lost. Under i.i.d.\ loss, dropped packets are spatially scattered at random. Under burst loss, consecutive packets are dropped together, producing a contiguous missing region.}
  \label{fig:pkloss_schematic}
\end{figure}

To assess robustness beyond i.i.d.\ packet erasures, we further evaluate a correlated burst-loss setting on DAIR-V2X under 90\% packet loss. Figure~\ref{fig:pkloss_schematic} illustrates the difference between the two loss models: i.i.d.\ loss drops spatial packets independently and uniformly, whereas burst loss removes a contiguous rectangular region of packets, producing a large hole in the received feature map. For each transmitted communication feature map, we first sample a coarse $8 \times 16$ Bernoulli keep mask with retention rate $p = 0.1$, then upsample it with nearest-neighbor interpolation to the native communication resolution. This produces contiguous missing regions while keeping the burst pattern normalized across methods with different communication resolutions. The ego feature is always kept, and the same per-sample mask is reused across methods for reproducibility and fair comparison.

\begin{table*}[!t]
  \centering
  \caption{Burst-loss results on DAIR-V2X under 90\% packet loss. F-Cooper is a dense BEV feature fusion baseline with 8\,MB communication, while CodeFilling and \methodname both operate at 128\,KB per transmitted collaborator frame.}
  \label{tab:burst_loss}
  \resizebox{\textwidth}{!}{
  \begin{tabular}{l|ccc|ccc|ccc|ccc|c}
      \toprule
      \multirow{2}{*}{\textbf{Models}} & \multicolumn{3}{c|}{\textbf{Short}} & \multicolumn{3}{c|}{\textbf{Middle}} & \multicolumn{3}{c|}{\textbf{Long}} & \multicolumn{3}{c|}{\textbf{mAP (\%)}} & \multirow{2}{*}{\textbf{Comm}} \\
      \cmidrule(lr){2-4} \cmidrule(lr){5-7} \cmidrule(lr){8-10} \cmidrule(lr){11-13}
      & 0.3 & 0.5 & 0.7 & 0.3 & 0.5 & 0.7 & 0.3 & 0.5 & 0.7 & 0.3 & 0.5 & 0.7 & \\
      \midrule
      F-Cooper & 79.15 & 76.03 & 65.84 & \textbf{76.34} & \textbf{71.05} & \textbf{56.56} & \textbf{55.91} & \textbf{50.68} & \textbf{37.78} & \textbf{69.88} & \textbf{65.08} & \textbf{52.12} & 8MB \\
      CodeFilling & 78.02 & 76.29 & 66.81 & 71.36 & 68.42 & 55.72 & 50.26 & 47.19 & 36.63 & 65.36 & 62.62 & 51.48 & 128KB \\
      \rowcolor{oursrow}
      \methodname & \textbf{79.96} & \textbf{77.53} & \textbf{66.97} & 74.15 & 70.22 & 56.38 & 52.70 & 48.75 & 36.99 & 67.83 & 64.21 & 51.93 & 128KB \\
      \bottomrule
  \end{tabular}
  }
\end{table*}

Table~\ref{tab:burst_loss} shows that the relative trend between low-bandwidth methods is preserved under burst loss. At the same 128\,KB budget, \methodname outperforms CodeFilling across all overall mAP thresholds (+2.47/+1.59/+0.45 at IoU 0.3/0.5/0.7) and achieves the best short-range AP at all three thresholds. F-Cooper still attains the highest overall mAP in this setting, which is consistent with its much larger 8\,MB communication budget and the larger amount of information it retains under correlated loss. Thus, burst loss is a stricter low-bandwidth reconstruction setting, but the method-level trend matches the main budget-matched comparison.

\subsection{Bandwidth--Robustness Analysis}
\label{app:bandwidth_robustness}

A recurring observation in our experiments is that dense BEV feature fusion baselines (8\,MB per transmitted collaborator frame) remain comparatively stable under packet loss. This behavior is closely tied to absolute information retention: under the same retention rate, a much larger communication budget leaves substantially more surviving information after transmission.

\textbf{Information retention under packet loss.}
Let $B$ denote the per transmitted collaborator frame communication budget and $p$ the packet retention rate. The absolute information surviving transmission is $B \cdot p$. Table~\ref{tab:info_retention} compares the retained information across representative bandwidth regimes at 90\% packet loss ($p = 0.1$):

\begin{table}[!htbp]
  \centering
  \caption{Absolute information retained under 90\% packet loss at different bandwidth levels.}
  \label{tab:info_retention}
  \resizebox{\linewidth}{!}{%
  \begin{tabular}{lccc}
      \toprule
      \textbf{Method class} & \textbf{Bandwidth $B$} & \textbf{Retained ($B \times 10\%$)} & \textbf{Rel. retained} \\
      \midrule
      Dense BEV feature fusion (e.g., F-Cooper) & 8\,MB & 819.2\,KB & $64\times$ \\
      Medium compression ($\times$2) & 2\,MB & 204.8\,KB & $16\times$ \\
      Our default ($\times$8) & 128\,KB & 12.8\,KB & $1\times$ \\
      Aggressive ($\times$16) & 32\,KB & 3.2\,KB & $0.25\times$ \\
      \bottomrule
  \end{tabular}
  }
\end{table}

At 90\% loss, a dense 8\,MB method still retains 819.2\,KB, which is $64\times$ more than the retained information of a 128\,KB method at the same loss rate and $6.4\times$ more than a complete 128\,KB transmission. Dense BEV feature fusion baselines under severe loss therefore continue to operate with substantially more raw information than low-bandwidth methods. Table~\ref{tab:budget_equalized_dense} complements the native-setting comparison by re-evaluating these dense BEV feature fusion baselines under the same 128\,KB budget as \methodname.

\begin{table*}[!t]
  \centering
  \caption{Budget-matched comparison on DAIR-V2X under 90\% packet loss. Dense BEV feature fusion baselines are contrasted between their native dense BEV setting (left) and the protocol-matched 128\,KB setting (right). \methodname is shown once as the native 128\,KB reference.}
  \label{tab:budget_equalized_dense}
  \setlength{\tabcolsep}{5.2pt}
  \renewcommand{\arraystretch}{1.12}
  \resizebox{\textwidth}{!}{
  \begin{tabular}{l|cccc|ccccc}
      \toprule
      \rule{0pt}{2.7ex}\textbf{Method} & \multicolumn{4}{c|}{\cellcolor{grouphdr}\textbf{Dense Native (8\,MB)}} & \multicolumn{5}{c}{\cellcolor{matchhdr}\textbf{Budget-Matched (128\,KB)}} \\
      \rowcolor{grouphdr}
      & mAP@0.3 & mAP@0.5 & mAP@0.7 & FPS & mAP@0.3 & mAP@0.5 & mAP@0.7 & FPS & $\Delta$ mAP@0.5 \\
      \midrule
      Attn & 66.87 & 66.20 & 52.56 & 6.67 & 62.99 & 60.25 & 48.65 & 6.91 & \cellcolor{losscell}-5.95 \\
      F-Cooper & 67.89 & 65.90 & 52.87 & 5.39 & 63.24 & 61.67 & 48.67 & 9.30 & \cellcolor{losscell}-4.23 \\
      CoBEVT & 68.83 & 63.55 & 44.47 & 4.92 & 67.73 & 62.85 & 52.99 & 5.30 & \cellcolor{losscell}-0.70 \\
      V2X-VIT & 68.78 & 64.33 & 50.97 & 4.70 & 64.69 & 60.53 & 49.21 & 5.32 & \cellcolor{losscell}-3.80 \\
      HEAL & 64.43 & 61.68 & 50.50 & 7.24 & 60.81 & 58.08 & 47.25 & 10.00 & \cellcolor{losscell}-3.60 \\
      \midrule
      \rowcolor{oursrow}
      \methodname & \multicolumn{4}{c|}{\textemdash} & 69.35 & 66.90 & 53.37 & 6.04 & \textemdash \\
      \bottomrule
  \end{tabular}
  }
  \vspace{0.15em}
  {\raggedright\footnotesize Dense-native values are copied from Table~\ref{tab:dair_v2x_comparison} under 90\% packet loss. The 128\,KB block reports equalized-budget re-evaluation, and $\Delta$ mAP@0.5 is defined as mAP@0.5(128\,KB) $-$ mAP@0.5(native dense setting). \methodname is shown only in the 128\,KB block because its native communication budget is already 128\,KB.\par}
\end{table*}

\textbf{Connecting to experimental results.}
This analysis helps interpret the trends in Table~\ref{tab:performance}. Dense BEV feature fusion baselines such as Attn, F-Cooper, and HEAL degrade less sharply from no loss to 90\% loss in part because they retain 819.2\,KB of absolute information, whereas CodeFilling operates under the same 128\,KB budget as \methodname. Table~\ref{tab:budget_equalized_dense} makes this comparison more direct: after enforcing the same 128\,KB budget, all five dense BEV feature fusion baselines decrease in mAP@0.5, with Attn decreasing by 5.95 points and even the most stable CoBEVT still decreasing by 0.70 points. Under this matched budget, \methodname achieves the best mAP among the 128\,KB entries and exceeds the strongest dense BEV feature fusion baseline at 128\,KB (CoBEVT) by 1.62/4.05/0.38 points at mAP@0.3/0.5/0.7.

\FloatBarrier

\section{Additional Ablations and Qualitative Results}
\label{app:additional_experiments}

This section expands the ablations summarized in Sec.~\ref{sec:ablation}. Unless noted otherwise, these results use DAIR-V2X and the same backbone, training schedule, and packet-loss protocol as the main experiments.

\subsection{Reconstructor Conditioning Inputs}

\begin{table}[!htbp]
  \centering
  \caption{Ablation of reconstructor conditioning inputs on DAIR-V2X under complete communication (no packet loss). All variants reconstruct from the same latent-derived prior; the table only removes ego-BEV, transformation, and direct latent-conditioning pathways inside the reconstructor.}
  \label{tab:ablation}
  \begin{tabular}{ccc|ccc}
      \toprule
      \multicolumn{3}{c|}{\textbf{Condition}} & \multicolumn{3}{c}{\textbf{mAP (\%)}} \\
      \cmidrule(r){1-3} \cmidrule(l){4-6}
      Ego feat. & Trans.\ mat. & Latent feat. & mAP@0.3 & mAP@0.5 & mAP@0.7 \\
      \midrule
                 &            &            & 67.28 & 63.71 & 52.01 \\
      \checkmark &            &            & 78.90 & 74.15 & 55.77 \\
      \checkmark & \checkmark &            & 79.69 & 75.34 & 56.56 \\
                 &            & \checkmark & 67.30 & 63.75 & 52.04 \\
                 & \checkmark & \checkmark & 68.30 & 64.74 & 53.04 \\
      \checkmark &            & \checkmark & 79.52 & 75.29 & 56.72 \\
      \rowcolor{oursrow}
      \checkmark & \checkmark & \checkmark & \textbf{81.03} & \textbf{76.16} & \textbf{57.74} \\
      \bottomrule
  \end{tabular}
\end{table}

Table~\ref{tab:ablation} shows that ego-side BEV context is the most important reconstructor condition. Since every row still reconstructs from the same latent-derived prior, the first row is not a no-collaboration setting; it tests whether the DiT can refine the prior without additional scene context. Adding aligned ego BEV gives the largest gain, indicating that the ego feature supplies geometric anchors in overlapping regions and helps the reconstructor distinguish plausible object structure from latent artifacts.

The transformation matrix and direct latent branch provide smaller but complementary refinements. The transformation contributes most when ego BEV is present because it tells the reconstructor how the ego context should be interpreted in the collaborator frame. The direct latent branch alone gives little gain because most collaborator evidence has already been decoded into the LPD prior, but combining all three conditions gives the highest mAP, suggesting that the branch still preserves residual compact cues that are useful after ego-guided refinement.

\subsection{Latent Prior Decoder Ablation}

\begin{figure}[!htbp]
  \centering
  \includegraphics[width=0.92\linewidth]{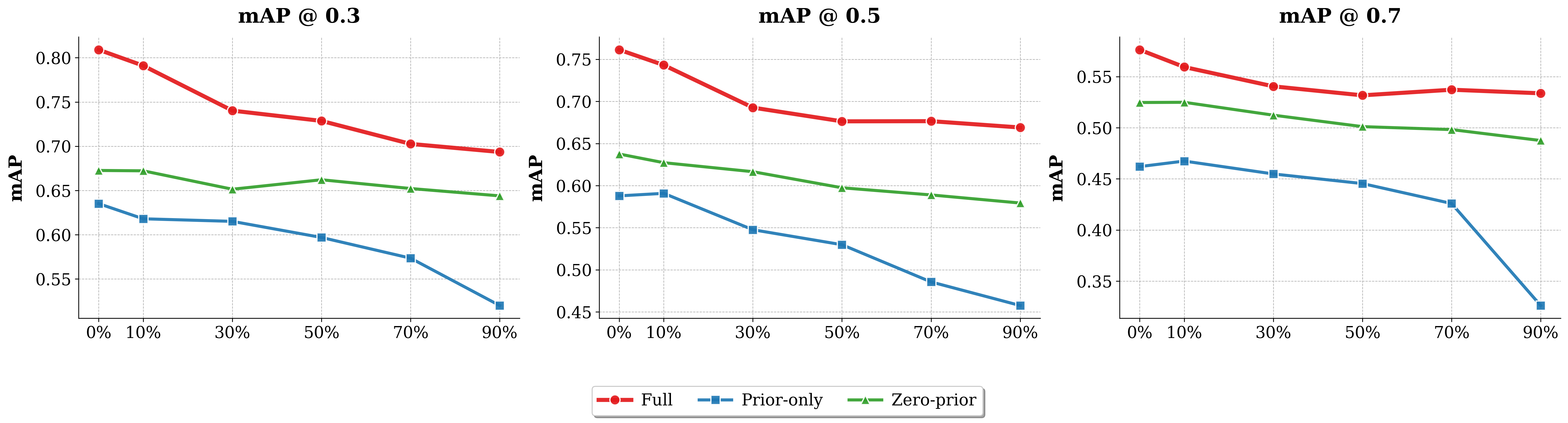}
  \caption{Ablation of the Latent Prior Decoder (LPD) on DAIR-V2X. We compare the full model, prior-only (using the LPD output directly and skipping the denoising reconstructor), and zero-prior (replacing the LPD initialization with an all-zero BEV prior before denoising) under varying packet-loss rates.}
  \label{ablation_lpd_fig}
\end{figure}

Fig.~\ref{ablation_lpd_fig} separates three roles of the reconstruction module. The full model uses the LPD output as a coarse initialization and then applies the noise-conditioned reconstructor. The prior-only variant removes this refinement stage and sends the LPD output directly to fusion, whereas the zero-prior variant keeps the denoising reconstructor but removes the latent-derived initialization by starting from an all-zero BEV prior. The full model remains consistently above both alternatives across packet-loss rates and IoU thresholds, showing that the two stages address different parts of the recovery problem.

The prior-only variant degrades most rapidly as packet loss increases. This is expected because the LPD is a deterministic decoder from the surviving latent packets, so when large spatial regions are missing it can only propagate coarse context from the received locations and cannot actively resolve ambiguous object evidence. The zero-prior variant remains functional because the denoiser can still use ego context, direct latent conditioning, and learned scene regularities; however, without an object-aligned LPD initialization, it has to infer more of the collaborator feature from weak context. Thus the LPD is not a redundant component: it provides a spatially aligned starting point, while diffusion-style denoising supplies the adaptive refinement needed under packet loss.

\subsection{Spatial Downsampling Ratio}

\begin{table}[!htbp]
  \centering
  \caption{Effect of spatial downsampling ratio on DAIR-V2X.}
  \label{tab:downsample}
  \begin{tabular}{ccccc}
      \toprule
      Downsample & mAP@0.3 & mAP@0.5 & mAP@0.7 & Comm \\
      \midrule
      $\times$2  & 82.55 & 77.64 & 59.42 & 2MB   \\
      $\times$4  & 81.52 & 76.79 & 58.52 & 512KB \\
      \rowcolor{oursrow}
      $\times$8  & 81.03 & 76.16 & 57.74 & 128KB \\
      $\times$16 & 79.58 & 73.51 & 54.52 & 32KB  \\
      \bottomrule
  \end{tabular}
\end{table}

Table~\ref{tab:downsample} shows that the spatial ratio controls both bandwidth and the amount of local structure retained in the latent prior. Moving from $\times$2 to $\times$8 reduces communication from 2\,MB to 128\,KB, a $16\times$ reduction, while the loss is only 1.52/1.48/1.68 mAP points at IoU 0.3/0.5/0.7. The default $\times$8 setting therefore sits near the knee of the accuracy--bandwidth curve: it removes most of the communication load while retaining enough spatial detail for the LPD and denoiser to reconstruct object-level BEV evidence.

The $\times$16 setting is more aggressive and begins to under-specify the reconstruction target. Its mAP@0.7 drop is larger than its mAP@0.3 drop, suggesting that coarse object presence can still be recovered but precise localization and shape are harder to preserve. This matches the role of the latent prior: when each latent cell summarizes too large a BEV region, the prior becomes smoother and less object-aligned, so the denoiser has to rely more heavily on ego context and learned priors rather than transmitted collaborator evidence.

\subsection{Noise-Conditioned Training and Inference Timestep}

\begin{table}[!htbp]
  \centering
  \caption{Effect of noise-conditioned training on DAIR-V2X under 90\% packet loss. The no-noise reconstructor keeps the same latent-derived prior and conditioning inputs but feeds the prior directly to the reconstructor at $t=0$, without forward-noise perturbation.}
  \label{tab:nonoise_compare}
  \begin{tabular}{lccc}
      \toprule
      Method & mAP@0.3 & mAP@0.5 & mAP@0.7 \\
      \midrule
      No-noise reconstructor & 62.19 & 59.68 & 49.14 \\
      \rowcolor{oursrow}
      \methodname ($t_{\text{inf}} = 700$) & \textbf{69.35} & \textbf{66.90} & \textbf{53.37} \\
      \bottomrule
  \end{tabular}
\end{table}

\begin{figure}[!htbp]
  \centering
  \includegraphics[width=0.92\linewidth]{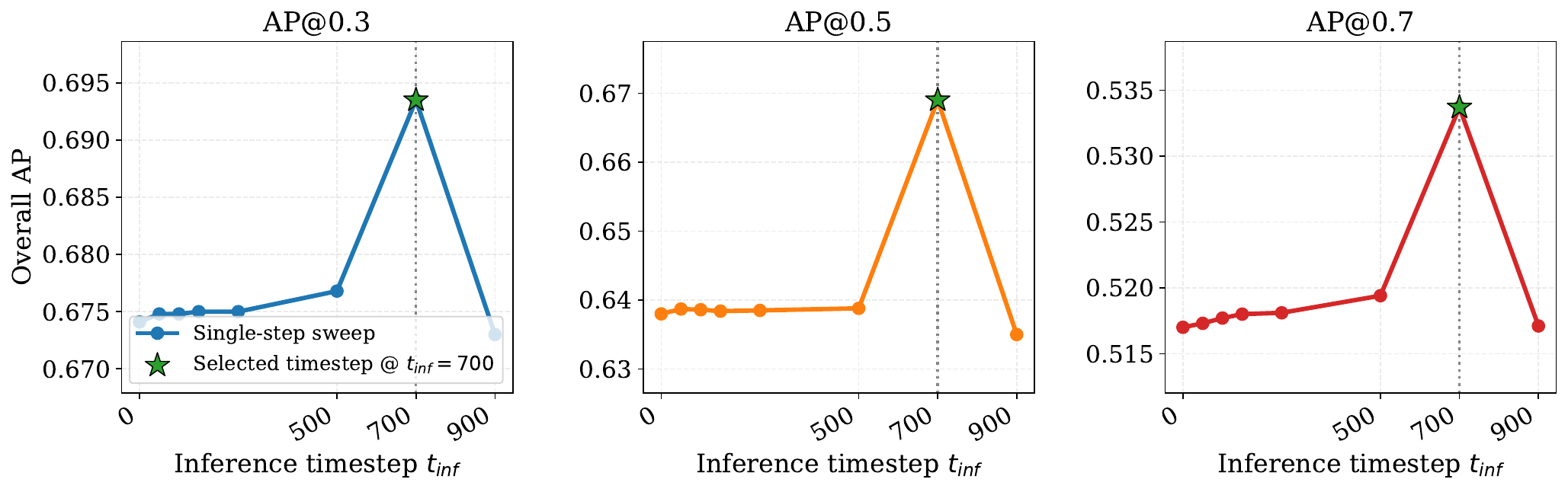}
  \caption{Sensitivity to the single-step inference timestep $t_{\text{inf}}$ on DAIR-V2X under 90\% packet loss. Solid curves show sweep results from a fixed checkpoint across different $t_{\text{inf}}$ values, and the star marks the selected timestep $t_{\text{inf}} = 700$. The trend is broadly stable across timesteps, indicating that the single-step denoiser is not highly sensitive to the exact inference timestep within a wide operating range.}
  \label{fig:tinf_sweep}
\end{figure}

Table~\ref{tab:nonoise_compare} isolates the value of noise-conditioned training under the main 90\% packet-loss setting. The no-noise reconstructor is different from both prior-only and zero-prior in Fig.~\ref{ablation_lpd_fig}: it still uses the LPD prior and a learned reconstructor, but it removes the diffusion-style perturbation by feeding the prior directly at $t=0$. Without this perturbation, the module is closer to an ordinary deterministic reconstructor. It learns to map a received latent prior to a target BEV feature, but it is not explicitly trained to calibrate how much of the prior should be trusted when large parts of the message are missing. Noise-conditioned training exposes the model to a continuum of corrupted priors, making the denoiser learn when to preserve the prior, when to use ego context, and when to fill missing collaborator evidence from learned spatial regularities.

Fig.~\ref{fig:tinf_sweep} further explains why the diffusion-style formulation is useful even with a single inference step. Low inference timesteps keep the input close to the LPD prior and may leave packet-loss artifacts under-corrected; very high timesteps inject more uncertainty and can weaken transmitted evidence. The selected $t_{\text{inf}}=700$ performs best in this sweep, while neighboring settings remain broadly stable, indicating that the method is not relying on a narrow condition-specific timestep.

\subsection{Qualitative Reconstruction}

\begin{figure}[!htbp]
  \centering
  \includegraphics[width=0.92\linewidth]{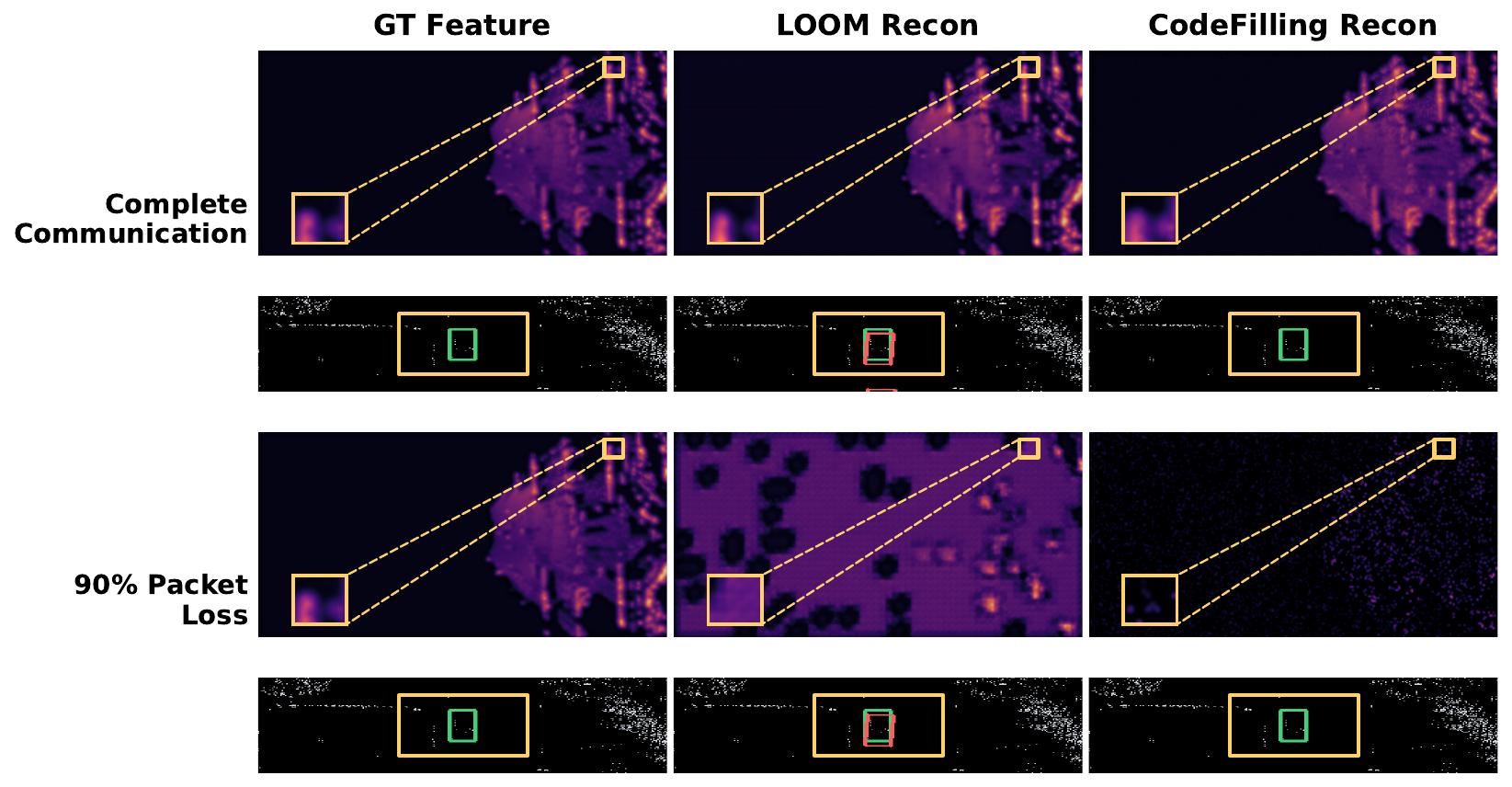}
  \caption{Detection-guided qualitative comparison between \methodname and CodeFilling on a fixed DAIR-V2X sample. The same target region is highlighted across complete communication and 90\% packet loss. Under severe loss, \methodname reconstructs a clearer object-centric BEV response and preserves the correct downstream detection, while CodeFilling misses the same target.}
  \label{recon_vis_fig}
\end{figure}

Fig.~\ref{recon_vis_fig} provides a qualitative example that is consistent with the quantitative packet-loss results. Under complete communication, both reconstructions retain a visible response around the highlighted target, so the difference is not simply due to the complete-communication setting. Under 90\% packet loss, CodeFilling produces a weaker and noisier response in the target region, and the downstream detector misses the object. In contrast, \methodname reconstructs a more coherent object-centric activation that aligns with the ground-truth feature and preserves the detection.

This visualization supports the high-IoU trend in Table~\ref{tab:performance}. Packet loss removes spatially localized evidence, and simple feature filling can leave either holes or diffuse artifacts that are insufficient for precise box regression. The LPD supplies a coarse object-aligned prior from surviving packets, while the denoising reconstructor sharpens that prior using ego context; the result is not just smoother BEV texture, but a feature pattern that remains useful to the detection head.

\begin{figure}[!htbp]
  \centering
  \includegraphics[width=\linewidth]{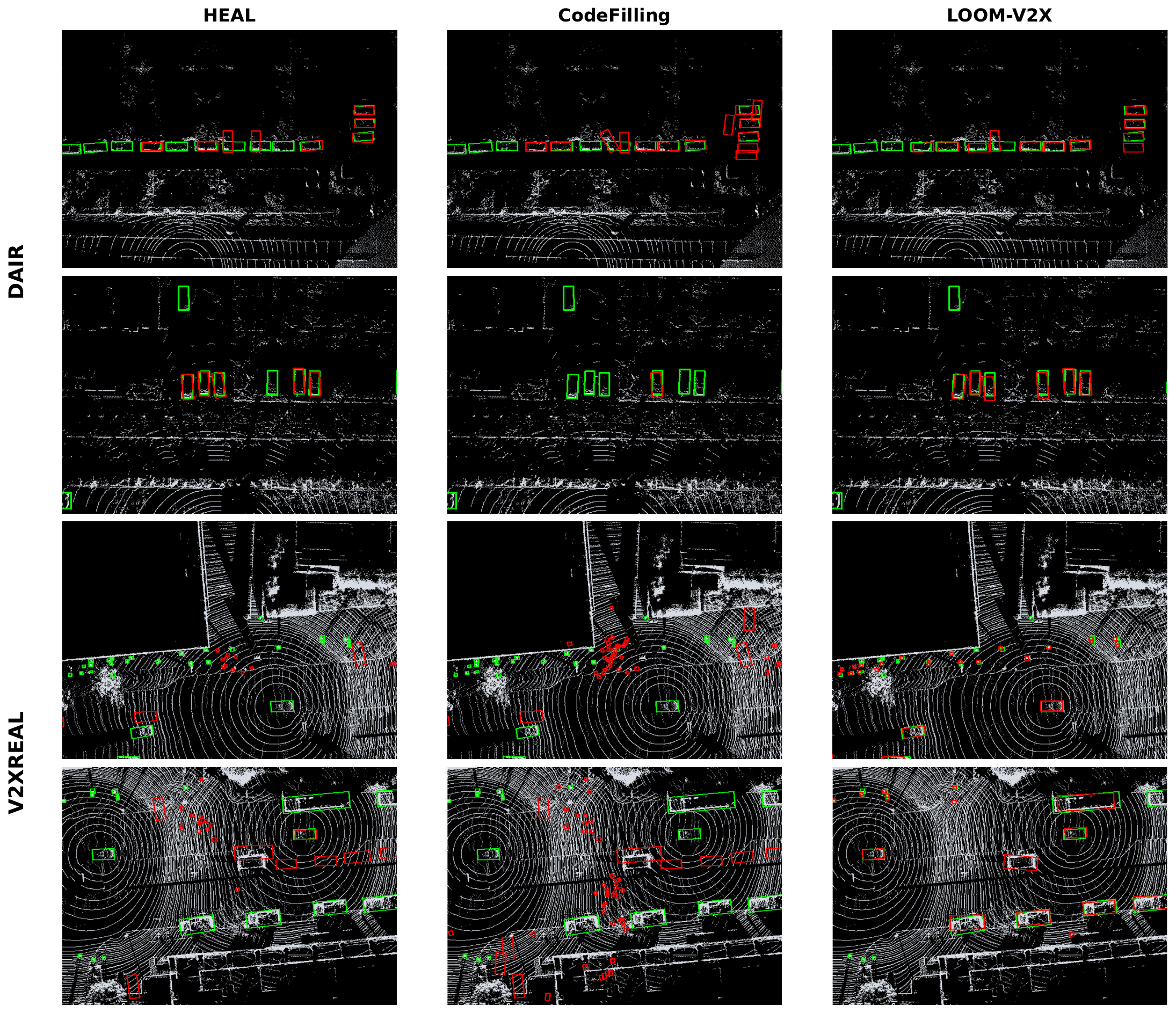}
  \caption{Visualization of detection results on DAIR-V2X and V2XREAL under 90\% packet loss. Columns compare HEAL, CodeFilling, and \methodname on the same samples. The top two rows are DAIR-V2X examples, and the bottom two rows are V2XREAL examples. Red boxes denote predictions and green boxes denote ground-truth boxes.}
  \label{fig:qual_compare_detection}
\end{figure}

Fig.~\ref{fig:qual_compare_detection} complements the feature-level visualization by presenting final detection outputs across multiple samples and datasets under 90\% packet loss. In these examples, HEAL and CodeFilling produce more false positives or missed objects when collaborator evidence is sparse, especially in crowded regions. In contrast, \methodname produces detections that are more consistently aligned with the ground-truth layout.

\FloatBarrier

\section{Clarifications on Design Choices}
\label{app:design_clarifications}

This section provides brief technical clarifications for two design choices used in \methodname.

\subsection{Single-Step Reconstruction vs.\ Iterative Sampling}
\label{app:single_step}

\methodname uses a single DDIM-style step at inference because reconstruction starts from the LPD prior BEV map $X_i^{\text{prior}}$ rather than from an uninformative initialization. The denoiser is trained with an $x_0$-prediction objective over noisy versions of this prior BEV map. At inference, Gaussian noise perturbs the prior to the fixed level $t_{\text{inf}}=700$, and one forward pass predicts the reconstructed collaborator BEV. Fig.~\ref{fig:tinf_sweep} is consistent with this interpretation: performance remains broadly stable across a wide range of $t_{\text{inf}}$ values, indicating that the reconstruction step is not highly sensitive to the exact inference timestep.

\subsection{Role of the Ego BEV Condition}
\label{app:ego_condition}

The ego vehicle's own BEV feature serves as complementary spatial context for reconstruction. For collaborator $i$, the ego feature is first warped into the collaborator coordinate frame as $X_0^{\text{aligned}} = \mathcal{W}(X_0, T_{0 \to i})$, matching the reconstruction formulation in Sec.~\ref{problem_formulation}. This aligned ego feature provides geometric anchors in overlapping regions and broader scene context around missing collaborator features. After reconstruction, the collaborator feature is warped back to the ego frame before fusion. Table~\ref{tab:ablation} supports this role: removing the ego BEV condition produces the largest performance drop among the conditioning ablations.

\section{Experimental Scope and Extensions}
\label{app:limitations}

The current evaluation focuses on real-world two-party V2X benchmarks, DAIR-V2X and V2XREAL, which provide a controlled setting for isolating lossy low-bandwidth communication from other multi-agent factors. Extending the same reconstruction protocol to larger cooperative graphs is a natural next step. Our default packet-loss model uses i.i.d.\ spatial erasures for reproducibility, and Appendix~\ref{app:burst_loss} adds a correlated burst-loss evaluation to cover a stronger spatial corruption pattern. Future extensions can incorporate richer channel simulators that jointly model latency, congestion, retransmission, and temporal burst dynamics. On the deployment side, lightweight denoisers, distillation, or adaptive invocation are practical directions for improving inference efficiency without changing the communication protocol.

\end{document}